\documentclass{article}

 \usepackage[main, final]{neurips_2026}
 \usepackage{microtype}
\usepackage{graphicx}
\usepackage{subcaption}
\usepackage{booktabs}
\usepackage{multirow} 
\usepackage{xcolor}   
\usepackage{amsmath}
\usepackage{amssymb}
\usepackage{mathtools}
\usepackage{amsthm}
\definecolor{mypink}{RGB}{255,105,180}
\usepackage{hyperref}
\usepackage[capitalize,noabbrev]{cleveref}
\usepackage[textsize=tiny]{todonotes}

\usepackage[utf8]{inputenc} % allow utf-8 input
\usepackage[T1]{fontenc}    % use 8-bit T1 fonts
\usepackage{hyperref}       % hyperlinks
\usepackage{url}            % simple URL typesetting
\usepackage{booktabs}       % professional-quality tables
\usepackage{amsfonts}       % blackboard math symbols
\usepackage{nicefrac}       % compact symbols for 1/2, etc.
\usepackage{microtype}      % microtypography
\usepackage{xcolor}         % colors

\title{DreamWorld: Unified World Modeling in Video Generation}

\author{%
  Boming Tan$^{1*}$, Xiangdong Zhang$^{2*}$, Ning Liao$^{2}$\thanks{Equal contribution. $\dagger$: Shaofeng Zhang is the corresponding author. This work was supported by the Fundamental and Interdisciplinary Disciplines Breakthrough Plan of the Ministry of Education of China and NSFC Youth Project 62606504.}, Jingtao Zhang$^{3}$, Yuqing Zhang$^{1}$, \\
  \textbf{Xiaoqiu Zhong$^{2}$, Xiaosong Jia$^{4}$, Xue Yang$^{2}$, Yanyong Zhang$^{1}$, Shaofeng Zhang$^{1\dagger}$} \\
  $^{1}$University of Science and Technology of China, $^{2}$Shanghai Jiao Tong University,\\
  $^3$Uber, $^4$Fudan University\\
  \texttt{bomingtan@foxmail.com, sfzhang@ustc.edu.cn}\\
}

\begin{document}

\maketitle

\begin{abstract}
Despite impressive progress in video generation, existing models remain limited to surface-level plausibility, lacking a coherent and unified understanding of the world. Prior approaches typically incorporate only a single form of world-related knowledge or rely on rigid alignment strategies to introduce additional knowledge. 
However, aligning the single world knowledge is insufficient to constitute a world model that requires jointly modeling multiple heterogeneous dimensions (e.g., physical commonsense, 3D and temporal consistency). 
To address this limitation, we introduce \textbf{DreamWorld}, a unified framework that integrates complementary world knowledge into video generators via a \textbf{Joint World Modeling Paradigm}, jointly predicting video pixels and features from foundation models to capture temporal dynamics, spatial geometry, and semantic consistency. 
However, naively optimizing these heterogeneous objectives can lead to visual instability and temporal flickering. To mitigate this issue, we propose \textit{Consistent Constraint Annealing (CCA)} to progressively regulate world-level constraints during training, and \textit{Multi-Source Inner-Guidance} to enforce learned world priors at inference. Extensive evaluations show that DreamWorld improves world consistency, outperforming Wan2.1 by 2.26 points on VBench. Code is available at \href{https://github.com/ABU121111/DreamWorld}{\textcolor{mypink}{\textbf{Github}}}.
\end{abstract}
\begin{figure}[h]
    \centering
    \includegraphics[width=\linewidth]{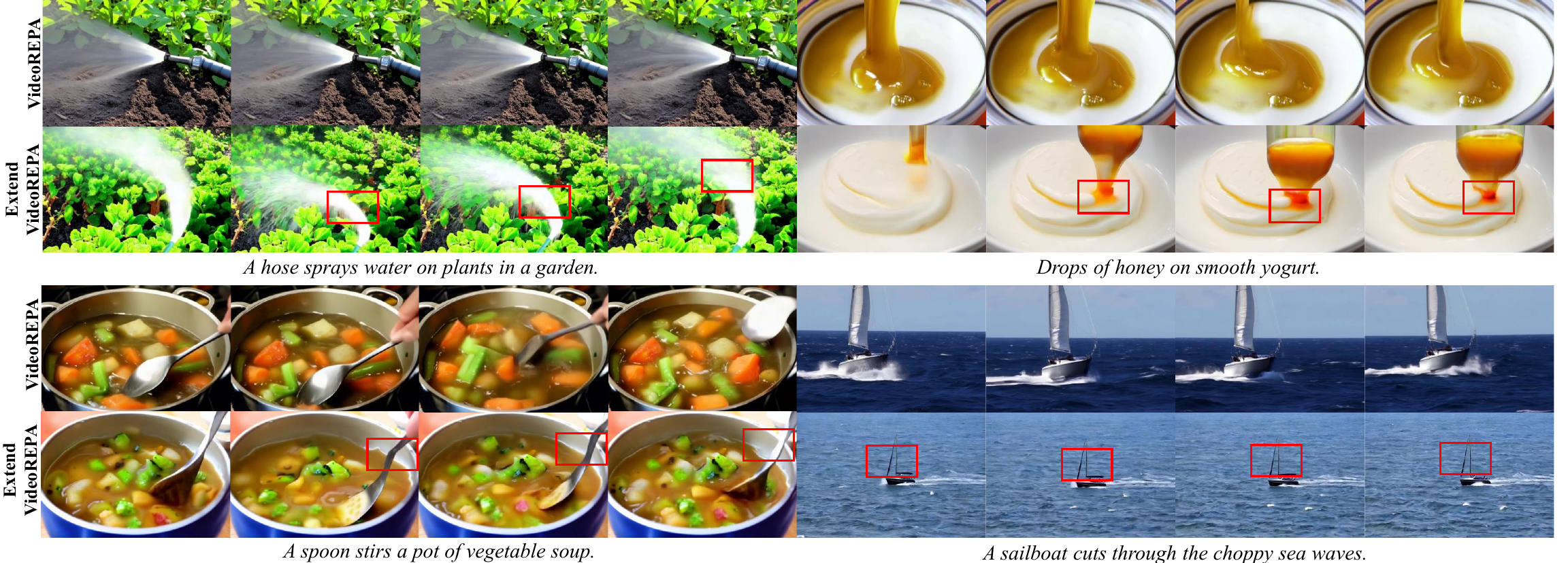}
    % \vspace{-10pt}
    \caption{Limitations of extending VideoREPA to multi-source knowledge, leading to structural implausibility and unnatural distortion. The physics score(PC) on the Videophy benchmark dropped from \textbf{29.7} to \textbf{24.1}.}
    \vspace{-15pt}
    \label{fig:motivation}
\end{figure}

\section{Introduction}
\label{sec:intro}

The landscape of text-to-video (T2V) generation has been transformed by scalable diffusion transformers~\cite{peebles2023scalablediffusionmodelstransformers, ma2025lattelatentdiffusiontransformer}, pushing the field beyond short visual clips toward the ambition of general-purpose {world models}~\cite{modify, videoworldsimulators2024}. Although state-of-the-art systems such as Lumiere~\cite{bartal2024lumierespacetimediffusionmodel}, OpenSora~\cite{zheng2024opensorademocratizingefficientvideo}, and Wan2.1~\cite{wan2025wanopenadvancedlargescale} achieve impressive cinematic fidelity through large-scale data and computation, a fundamental gap remains: these models largely function as visual generators rather than world simulators. Optimized primarily for pixel-level distribution matching~\cite{kang2025farvideogenerationworld,berg2025semanticworldmodels}, they fail to internalize structured and comprehensive world knowledge, which is reflected in their limited performance on world-centric benchmarks such as VBench 2.0~\cite{zheng2025vbench20advancingvideogeneration}.

To mitigate the gap between visual realism and world understanding, prior work~\cite{wu2025geometryforcingmarryingvideo,zhang2025videorepalearningphysicsvideo} has explored injecting external world knowledge into video generation models. A representative direction is Representation Alignment (REPA)~\cite{yu2025representationalignmentgenerationtraining}, which aligns generative models with pretrained experts to transfer structured priors. Building on this paradigm, VideoREPA~\cite{zhang2025videorepalearningphysicsvideo} introduces Token Relation Distillation (TRD) as a soft alignment strategy, demonstrating its effectiveness in distilling knowledge from a single expert model. However, constructing a holistic world model inherently requires the simultaneous integration of \textit{multiple heterogeneous knowledge sources}. Our empirical investigations reveal that naively extending the REPA to concurrently align with semantic, spatial, and dynamic experts leads to a multi-objective optimization dilemma. As illustrated in Figure~\ref{fig:motivation}, conflicting relational gradients from distinct teacher models induce optimization instability, suggesting that relational alignment is insufficient for the multi-source integration required by a unified world model (see Appendix~\ref{app:A}).

In response to these limitations, we propose \textbf{DreamWorld}, a unified framework designed to internalize comprehensive world knowledge through a \textbf{Joint World Modeling Paradigm}. Inspired by  VideoJAM~\cite{chefer2025videojamjointappearancemotionrepresentations}, DreamWorld extends standard video latents into a composite feature space, compelling the model to predict video pixels alongside a set of world features. Specifically, DreamWorld integrates temporal dynamics from Optical Flow, spatial geometry from VGGT~\cite{wang2025vggtvisualgeometrygrounded}, and semantic understanding from DINOv2~\cite{oquab2024dinov2learningrobustvisual}. 

However, the direct superposition of such heterogeneous optimization objectives frequently induces optimization instability and temporal flickering. To mitigate this, we propose \textit{Consistent Constraint Annealing (CCA)}, a decay mechanism that guarantees convergence by progressively modulating the influence of world knowledge, thereby ensuring high-fidelity visual generation while effectively assimilating world priors. Moreover, a \textit{Multi-Source Inner-Guidance} mechanism is incorporated at inference time, which leverages the model's own predicted knowledge features to steer the generation process, ensuring trajectories that strictly adhere to real-world laws. \textbf{Our main contributions are summarized as follows:}

\textbf{i)} We present \textbf{DreamWorld}, the first unified video generation framework to integrate multi-source world knowledge, including 3D Semantic consistency, Motion Temporal dynamics, and 2D Spatial geometry.

\textbf{ii)} We propose a novel training strategy, \textbf{Consistent Constraint Annealing (CCA)}, which harmonizes knowledge injection with visual quality, ensuring coherent and artifact-free generation.

\textbf{iii)} Extensive evaluations demonstrate that DreamWorld significantly outperforms baselines and VideoJAM, establishing a new standard for world models.

\section{Related Works}

\paragraph{Video Diffusion Models.}
The paradigm of video generation has been revolutionized by the adoption of Diffusion Transformer (DiT) architectures~\cite{peebles2023scalablediffusionmodelstransformers, ma2025lattelatentdiffusiontransformer} Following the scaling laws originally observed in language modeling~\cite{kaplan2020scalinglawsneurallanguage} and validated in video~\cite{tong2025thinkingvideovideogeneration}, recent state-of-the-art systems such as Wan2.1~\cite{wan2025wanopenadvancedlargescale} and Hunyuan-Video~\cite{kong2025hunyuanvideosystematicframeworklarge} have achieved photorealistic results by training on massive video-text corpora. To enhance training stability and inference efficiency beyond standard diffusion, Flow Matching~\cite{lipman2023flowmatchinggenerativemodeling} has emerged as a superior generative framework, as successfully implemented in LTX-Video~\cite{hacohen2026ltx2efficientjointaudiovisual} and Pyramid Flow~\cite{jin2025pyramidalflowmatchingefficient}. Furthermore, the open-source community has democratized access to high-fidelity generation through open-weights models like Mochi 1~\cite{genmo2024mochi} and CogVideoX~\cite{yang2025cogvideoxtexttovideodiffusionmodels}. Despite these achievements, pure diffusion-based approaches often struggle with real-world understanding, lacking the inherent capability for maintaining global consistency~\cite{qin2024worldsimbenchvideogenerationmodels}.

\paragraph{Representation Alignment.}
Representation Alignment (REPA)~\cite{yu2025repa} addresses the lack of structural awareness in pixel-space diffusion by injecting high-level semantic priors from pre-trained foundation models. Building on this, recent advancements have introduced more sophisticated alignment mechanisms~\cite{zheng2025diffusiontransformersrepresentationautoencoders,jiang2025representationcomponentneededdiffusion,lee2025aligningtextimagediffusion,li2025semanticspaceinterveneddiffusivealignmentvisual,zhao2025guidingcrossmodalrepresentationsmllm}. In the video domain, AlignVid~\cite{liu2025alignvidtrainingfreeattentionscaling}, VideoREPA~\cite{zhang2025videorepalearningphysicsvideo} and MoAlign~\cite{bhowmik2025moalignmotioncentricrepresentationalignment} extend this mechanism to spatio-temporal alignment, enforcing adherence to a coherent semantic layout in generated frames. Although these methods offer a pathway to more robust video generation, they typically focus on appearance consistency rather than dynamic causal logic.

\paragraph{World Modeling.}
The concept of World Models~\cite{modify} transcends video generation, aiming for a rigorous understanding of the environment's underlying laws to predict and simulate the world. Genie~\cite{bruce2024geniegenerativeinteractiveenvironments} and Genie 2~\cite{parkerholder2024genie2} instantiate this as an interactive simulator, learning a latent action space for controlling video rollouts. Alternatively, the Joint-Embedding Predictive Architecture (JEPA), exemplified by V-JEPA~\cite{bardes2024revisiting} and V-JEPA2~\cite{assran2025vjepa2selfsupervisedvideo}, utilizes non-generative predictive mechanisms for learning abstract world states. Frame-level~\cite{fuest2025maskflowdiscreteflowsflexible,song2025historyguidedvideodiffusion,chen2024diffusionforcingnexttokenprediction,wu2025videoworldmodelslongterm,po2025longcontextstatespacevideoworld} context mechanisms introduce frame-level context guidance by adding noise to context frames during training. Additionally, some methods~\cite{xiao2026worldmemlongtermconsistentworld} utilize 3D information to enhance spatial coherence. Hybrid approaches such as DriveWorld~\cite{min2024driveworld4dpretrainedscene} and UniWorld~\cite{lin2025uniworld} attempt the unification of generative decoding with state-space modeling. However, existing paradigms lack the capability to synergize knowledge from multiple heterogeneous expert models~\cite{wu2025geometryforcingmarryingvideo}. Bridging this gap, our work fuses complementary priors, resulting in a world model that surpasses implicit alignment methods in long-horizon consistency and realism.

\section{Method}
\label{sec:method}
\begin{figure*}[t]
    \centering
    \includegraphics[width=1\linewidth]{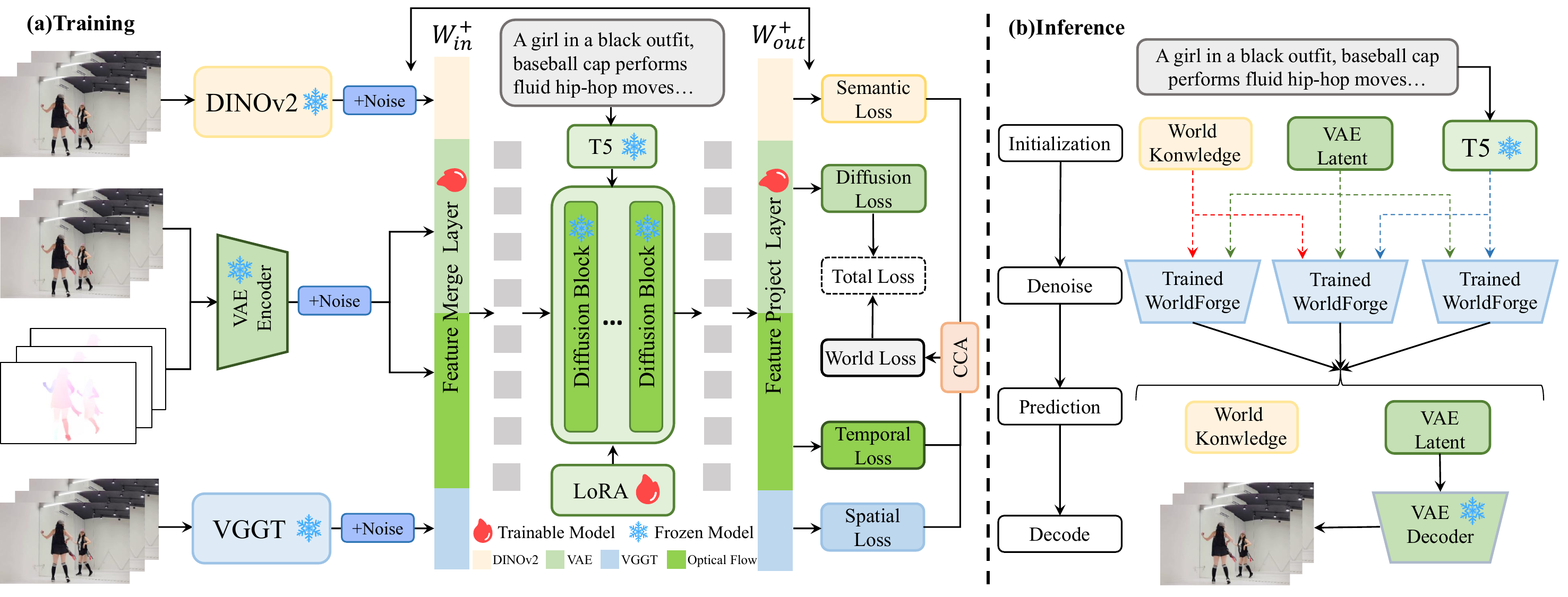} 
    \caption{\textbf{Overview of DreamWorld.} \textbf{(a) Training:} Expert models extract multimodal features, which are noised, concatenated, and fused via a linear layer $\mathbf{W_{in}^+}$. A subsequent layer $\mathbf{W_{out}^+}$ jointly predicts appearance and world knowledge, optimized by a World Loss with \textit{Consistency Constraint Annealing (CCA)} to ensure generation fidelity. \textbf{(b) Inference:} We introduce \textit{Multi-Source Inner-Guidance}, leveraging inherent noise features to direct the final video generation.}
    % \vspace{-15pt}
    \label{fig2:env}
\end{figure*}

\label{sec:method}

This section outlines the proposed framework. We first introduce the World Knowledge priors and the video diffusion backbone in Section~\ref{subsec:preliminaries}, followed by the unified preprocessing and alignment protocols in Section~\ref{subsec:preprocessing}. Section~\ref{subsec:architecture} then details the Joint World Knowledge Learning (see Figure~\ref{fig2:env}), optimized via the Consistent Constraint Annealing (CCA) strategy to balance physical constraints and generative fidelity. Finally, Section~\ref{subsec:guidance} presents the Multi-Source Inner-Guidance mechanism for controllable inference.

\vspace{-4pt}
\subsection{Preliminaries}
\vspace{-4pt}
\label{subsec:preliminaries}

\paragraph{Video Diffusion Transformers.}
Recent advances in text-to-video generation, exemplified by models such as Wan2.1~\cite{wan2025wanopenadvancedlargescale}, utilizes Flow Matching~\cite{lipman2023flowmatchinggenerativemodeling} transformers to model the transition from noise to video as a continuous-time process. The pipeline begins with a 3D Causal VAE that compresses the video $\mathbf{x}$ into latent representations $\mathbf{z}_0 = \mathcal{E}(\mathbf{x})$. The core generative mechanism learns to straighten the probability path between the target data $\mathbf{z}_0$ and the Gaussian prior $\mathbf{z}_1$. Defining the intermediate state as:
\begin{equation}
    \mathbf{z}_t = t\mathbf{z}_1 + (1-t)\mathbf{z}_0
\end{equation}
The model optimizes the velocity field $v_\theta$ by regressing against the constant flow $\mathbf{z}_1 - \mathbf{z}_0$:
\begin{equation}
    \mathcal{L}_{\text{flow}} = \mathbb{E}_{t, \mathbf{z}_0, \mathbf{z}_1} \left[ \| v_\theta(\mathbf{z}_t, t, \mathbf{c}) - (\mathbf{z}_1 - \mathbf{z}_0) \|^2 \right]
\end{equation}
where $\mathbf{c}$ represents the conditioning text embeddings. During inference, this learned velocity field guides the generation, allowing the model to recover the video latent $\hat{\mathbf{z}}_0$ by numerically solving the ordinary differential equation (ODE) from $t=1$ to $t=0$:
\begin{equation}
    \hat{\mathbf{z}}_0 = \mathbf{z}_1 + \int_{1}^{0} v_\theta(\mathbf{z}_\tau, \tau, \mathbf{c}) \, d\tau
\end{equation}
\paragraph{WorldKnowledge Priors.}
% We construct a composite feature space, $\mathcal{Z}_{\text{world}}$, that establishes the holistic understanding essential for a world model by unifying three fundamental dimensions of reality. Specifically, \textit{Optical Flow}~\cite{HORN1981185} encodes dense pixel-level trajectories for temporal dynamics. \textit{DINOv2}~\cite{oquab2024dinov2learningrobustvisual} provides robust semantic features to preserve objects following the prompt rules. Finally, \textit{VGGT}~\cite{wang2025vggtvisualgeometrygrounded} explicitly models spatial relationships in 2D geometric constraints.
We construct a composite feature space, $\mathcal{Z}_{\text{world}}$, encapsulating essential world knowledge by unifying the fundamental \textit{what}, \textit{where}, and \textit{how} of physical reality. Specifically, \textit{DINOv2}~\cite{oquab2024dinov2learningrobustvisual} extracts robust semantics to define what objects exist, ensuring appearance consistency. \textit{VGGT}~\cite{wang2025vggtvisualgeometrygrounded} models 2D geometric constraints to establish where entities reside spatially. Finally, \textit{Optical Flow}~\cite{HORN1981185} encodes dense pixel-level trajectories to govern how elements evolve over time. 

\vspace{-4pt}
\subsection{Preprocessing}
\vspace{-4pt}
\label{subsec:preprocessing}
We implement a unified preprocessing protocol comprising motion-to-RGB conversion, spatio-temporal alignment, and channel compression, thereby aligning heterogeneous priors with the diffusion backbone.
\paragraph{Motion Representation Transformation.}Since standard video encoders require RGB inputs, dense displacement fields $\mathbf{d} \in \mathbb{R}^{H \times W \times 2}$—computed through RAFT~\cite{teed2020raftrecurrentallpairsfield}—are mapped into a 3-channel space following the VideoJAM protocol~\cite{chefer2025videojamjointappearancemotionrepresentations}.
\begin{equation}
    m = \min\left(1, \frac{\sqrt{u^2 + v^2}}{\sigma \sqrt{H^2 + W^2}}\right), \alpha = \operatorname{arctan2}(v, u)
\end{equation}
where the normalized motion magnitude $m$ and motion direction $\alpha$ are computed from the optical flow components $(u, v)$ to modulate the pixel-wise intensity and hue, respectively, while $\sigma$ serves as a scaling factor that controls the sensitivity of the motion magnitude normalization. The resulting visualizations are subsequently sampled and encoded by a pre-trained 3D causal VAE~\cite{wan2025wanopenadvancedlargescale} as videos, yielding the compressed temporal latent $\mathbf{z}_{\text{temporal}}$. 

\paragraph{Unified Alignment and Integration.}
Raw video foundation model (VFM) outputs are spatially interpolated and temporally pooled to align with the latent dimensions. Given the aligned tensor $\mathbf{F} \in \mathbb{R}^{C \times T \times H \times W}$, we reconcile distributional discrepancies across semantic, temporal, and spatial features via global normalization:
\begin{equation}
    \hat{\mathbf{F}} = \frac{\mathbf{F} - \mu}{\sigma + \epsilon}
\end{equation}
where $\mu$ and $\sigma$ are the scalar mean and standard deviation across all dimensions, and $\epsilon = 10^{-6}$. Next, we apply PCA~\cite{kouzelis2026boostinggenerativeimagemodeling} to compress $\hat{\mathbf{F}}$, filtering redundant noise and distilling features into compact latents: $\mathbf{z}_{\text{temporal}}$, $\mathbf{z}_{\text{semantic}}$, and $\mathbf{z}_{\text{spatial}}$. Finally, we concatenate these priors along the channel dimension to form the consolidated world latent, efficiently conditioning the model without excessive computational overhead:
\begin{equation}
    \mathbf{Z}_{\text{world}} = [\mathbf{z}_{\text{temporal}}, \mathbf{z}_{\text{semantic}}, \mathbf{z}_{\text{spatial}}]
\end{equation}

\vspace{-4pt}
\subsection{Joint World Knowledge Learning}
\vspace{-4pt}
\label{subsec:architecture}
Rather than treating $\mathbf{Z}_{\text{world}}$ merely as a conditioning signal~\cite{jang2025multidimensionalpreferencealignmentconditioning} or directly aligned representation~\cite{yu2025repa}, we concatenate $\mathbf{Z}_{\text{world}}$ with the latent and embed the joint features into the diffusion block, enabling the learning of mutual information between visual appearance and underlying world knowledge.

% \paragraph{Joint World Knowledge Modeling.}
% The core architectural modification lies in the expansion of the linear projection layers. As illustrated in standard dual-stream protocols from VideoJAM, we extend the pre-trained input projection layer $\mathbf{W}_{in}$ to accommodate the concatenated input $\mathbf{z}_t = [\mathbf{z}_{vae}, \mathbf{Z}_{world}]_t$. The new projection layer $\mathbf{W}_{in}^+$ is formulated as:
\paragraph{Joint Feature Integration.}
The core architectural modification in our approach focuses on enabling \textbf{joint modeling of video appearance and high-level world knowledge} within a unified transformer framework. Concretely, we expand the linear projection layers to accommodate inputs from both the denoising stream and the world knowledge stream. We extend the pre-trained input projection layer $\mathbf{W}_{\text{in}}$ to accommodate the concatenated input
$\mathbf{z}_t = [\mathbf{z}_{\text{vae}}, \mathbf{Z}_{\text{world}}]_t$.
The resulting projection layer is defined as:
% The core architectural modification in our approach focuses on enabling \textbf{joint modeling of video appearance and high-level world knowledge} within a unified transformer framework. Concretely, we expand the linear projection layers to accommodate inputs from both the denoising stream and the world knowledge stream. Following the dual-stream design principles in VideoJAM, we extend the pre-trained input projection layer $\mathbf{W}_{in}$ to process the concatenated input vector $\mathbf{z}_t = [\mathbf{z}_{vae}, \mathbf{Z}_{world}]_t$,
% where $\mathbf{z}_{vae}$ represents the latent features extracted from video frames via a pre-trained VAE, and $\mathbf{Z}_{world}$ encodes structured knowledge about the environment or task-specific world state. The expanded input projection is formally defined as: 
\begin{equation}
    \mathbf{W}_{\text{in}}^+ = [\mathbf{W}_{\text{in}}, \mathbf{0}] \in \mathbb{R}^{(C_{\text{vae}} + C_{\text{world}}) \times D}
\end{equation}
The weights corresponding to $\mathbf{Z}_{\text{world}}$ are initialized to \textbf{zero}, ensuring that the initial model behavior exactly matches the original pre-trained Wan2.1. This design stabilizes training by preventing abrupt interference from high-dimensional world features, while allowing their influence to be gradually learned through optimization.
Symmetrically, the output projection is expanded to $\mathbf{W}_{\text{out}}^+ \in \mathbb{R}^{D \times (C_{\text{vae}} + C_{\text{world}})}$ to predict a joint velocity field, which is subsequently decomposed into modality-specific components:
% Crucially, the weights corresponding to $\mathbf{Z}_{world}$ are initialized to \textbf{zero}, which ensures that at the onset of training, the network's behavior is equivalent to the original pre-trained Wan2.1. This prevents the sudden influx of high-dimensional multi-modal features from overwhelming the pre-trained weights, thereby facilitating a progressive and stable assimilation process where the model gradually learns to incorporate the WorldKnowledge priors.
% Symmetrically, the output projection is expanded to $\mathbf{W}_{out}^+ \in \mathbb{R}^{D \times (C_{vae} + C_{world})}$ to predict the joint velocity field. The output $\mathbf{z}_{pred}$ is then split to recover individual signal components:
\begin{equation}
\begin{gathered}
    \mathbf{z}_{\text{pred}} =\hat{\mathbf{v}}(\mathbf{z}_t,y,t)= \mathbf{M}_{(\mathbf{z}_t \bullet \mathbf{W}_{\text{in}}^+)} \bullet \mathbf{W}_{\text{out}}^+ 
    \xrightarrow{\text{split}} [\hat{\mathbf{z}}_{\text{vae}}, \hat{\mathbf{z}}_{\text{temporal}}, \hat{\mathbf{z}}_{\text{semantic}}, \hat{\mathbf{z}}_{\text{spatial}}]
\end{gathered}
\end{equation}
where $\mathbf{M}$ denotes the attention blocks and $y$ represents the conditioning prompt.

% where \(\mathbf{M}\) denotes the attention blocks, and \(\hat{\mathbf{v}}_i\) denotes the predicted composite features, $y$ denotes the given prompts.

% \paragraph{Training objective.}
% The model optimization is driven by a Flow Matching objective, where the total loss $\mathcal{L}_{total}$ is formulated as a weighted sum of reconstruction objectives across different modalities. Formally, the total loss function is defined as:
\paragraph{Training Objective with Consistent Constraint Annealing.}
The model is optimized using a Flow Matching objective defined over the joint velocity field. The total loss is formulated as a weighted sum of modality-specific flow matching terms:
\begin{equation}
    \begin{gathered}
        \mathcal{L}_{\text{total}} = \mathcal{L}_{\text{vae}} + \lambda_{\text{temp}}(t) \mathcal{L}_{\text{temporal}}
             + \lambda_{\text{sem}}(t) \mathcal{L}_{\text{semantic}} + \lambda_{\text{spa}}(t) \mathcal{L}_{\text{spatial}}
    \end{gathered}
\end{equation}
% Specifically, each individual loss term $\mathcal{L}_{k}$, where $k \in \{vae, temporal, semantic, spatial\}$, quantifies the flow matching error between the predicted velocity component $\hat{\mathbf{z}}_k$ and the target conditional flow $\mathbf{u}_t^{(k)}$ constructed from the ground truth feature $\mathbf{z}_k$, expressed as the expectation over the time step $t$, data distribution, and Gaussian noise path:
% \begin{equation}
%     \mathcal{L}_{k} = \mathbb{E}_{t, \mathbf{z}_k} \left[ \left\| \hat{\mathbf{z}}_k(\mathbf{z}_t, y, t) - \mathbf{u}_t^{(k)}(\mathbf{z}_k) \right\|_2^2 \right]
% \end{equation}
% where $\mathbf{u}_t^{(k)}$ denotes the target velocity field pointing to the ground truth $\mathbf{z}_k$.
Specifically, each individual loss term $\mathcal{L}_{k}$, where $k \in \{vae, temporal, semantic, spatial\}$, quantifies the flow matching error by computing the Mean Squared Error (MSE) between the predicted velocity component $\hat{\mathbf{z}}_k$ and the target conditional flow $\mathbf{u}_t^{(k)}$, which denotes the target velocity field pointing to the ground truth feature $\mathbf{z}_k$.

However, a static weighting scheme often exacerbates the inherent tension between maintaining high-fidelity \textbf{generative quality} and enforcing effective \textbf{knowledge learning} from the introduced features~\cite{fu2019cyclicalannealingschedulesimple}. To reconcile this conflict, we introduce \textit{Consistent Constraint Annealing (CCA)}, Unlike static strategies, CCA gradually relaxes these weights to zero, prioritizing high-fidelity, artifact-free visual reconstruction in the final phase. Mathematically, the weight $\lambda(t)$ at current training step $t$ is formulated as:
\begin{equation}
    \lambda(t) = \lambda_{\text{base}} \cdot \frac{1}{2} \left[ 1 + \cos\left( \frac{t}{T_{\text{total}}} \pi \right) \right]
    \label{cca}
\end{equation}
where $\lambda_{\text{base}}=0.2$ denotes the initial constraint intensity and $T_{\text{total}}$ is the total training duration.

\vspace{-4pt}
\subsection{Multi-Source Inner-Guidance}
\vspace{-4pt}
\label{subsec:guidance}
To effectively guide video generation with predicted world knowledge during inference, we introduce Multi-Source Inner-Guidance. This approach extends standard classifier-free guidance~\cite{vincent2011connection,ho2022classifierfreediffusionguidance} from a typical two-branch (conditional versus unconditional) framework into differential guidance across multiple conditional subspaces. During inference, the joint latent $[\mathbf{z}_{\text{vae}}, \mathbf{Z}_{\text{world}}]$ is initialized from a standard Gaussian distribution. While training uses expert-derived targets for $\mathbf{Z}_{\text{world}}$, inference relies on self-predicted $\mathbf{Z}_{\text{world}}$ maintained along the denoising trajectory. Dynamically generating $\mathbf{Z}_{\text{world}}$ ensures these world priors remain synchronized and aligned with the intermediate state $\mathbf{z}_t$ at each timestep, which intrinsically mitigates train-test distribution shift and error accumulation caused by mismatched external predictors. Formulated within the Flow Matching velocity domain, the rectified velocity field $\mathbf{z}_{\text{pred}}$ is computed by aggregating the fully conditioned prediction with the differential updates from each masked feature subspace:
\begin{equation}
\begin{gathered}
\mathbf{z}_{\text{pred}} \leftarrow
(1 + w_{\text{txt}} + w_{\text{temp}} + w_{\text{sem}} + w_{\text{spa}})\,\hat{\mathbf{v}}(\mathbf{z}_t,y,t) \\
- \underbrace{w_{\text{txt}} \hat{\mathbf{v}}(\mathbf{z}_t, \emptyset, t)}_{\text{\scriptsize Text Guidance}}
- \underbrace{w_{\text{temp}} \hat{\mathbf{v}}(\mathbf{z}_t^{\neg \text{temp}}, y, t)}_{\text{\scriptsize Motion Guidance}}
- \underbrace{w_{\text{sem}} \hat{\mathbf{v}}(\mathbf{z}_t^{\neg \text{sem}}, y, t)}_{\text{\scriptsize Semantic Guidance}}
- \underbrace{w_{\text{spa}} \hat{\mathbf{v}}(\mathbf{z}_t^{\neg \text{spa}}, y, t)}_{\text{\scriptsize Spatial Guidance}}
\end{gathered}
\end{equation}
where $\mathbf{z}_{t}^{\neg k}$ denotes masking the corresponding channel in the World Knowledge tensor. Empirically, we prioritize prompt adherence with $w_{\text{txt}}=5$, while assigning moderate weights $w_{\text{temp}}=w_{\text{sem}}=w_{\text{spa}}=1$ to the priors. A detailed Inference Cost Analysis is provided in Appendix \ref{app:B}.

\begin{table*}[tb!]
    \centering
    \caption{\textbf{Quantitative comparison on VBench.}\textbf{FT.} denotes the fine-tuned version, and \textbf{Reimpl.} indicates our re-implementation of the method.}
    \label{tab:vbench}
    \setlength{\tabcolsep}{2pt} 
    \resizebox{\linewidth}{!}{%
        \begin{tabular}{l ccccc ccccc c cc c}
            \toprule
            \multirow{2}{*}[-2ex]{\textbf{\large Method}} & 
            \multicolumn{5}{c}{\textbf{\small Temporal}} & 
            \multicolumn{5}{c}{\textbf{\small Semantic}} & 
            \multicolumn{1}{c}{\textbf{\small Spatial}} & 
            \multicolumn{2}{c}{\textbf{\small Summary}} &
            \multirow{2}{*}[-2ex]{\textbf{\small \shortstack{Overall\\Score}}} \\
            
            \cmidrule(lr){2-6} \cmidrule(lr){7-11} \cmidrule(lr){12-12} \cmidrule(lr){13-14}
            
            & \textbf{\shortstack{Subject\\Consistency}} 
            & \textbf{\shortstack{Background\\Consistency}} 
            & \textbf{\shortstack{Temporal\\Flickering}} 
            & \textbf{\shortstack{Motion\\Smoothness}} 
            & \textbf{\shortstack{Dynamic\\Degree}} 
            & \textbf{\shortstack{Object\\Class}} 
            & \raisebox{1.2ex}{\textbf{\shortstack Color}}
            & \textbf{\shortstack{Human\\Action}} 
            & \textbf{\shortstack{Multiple\\Objects}}
            & \raisebox{1.2ex}{\textbf{Scene}} 
            & \raisebox{1.2ex}{\textbf{Relationship}}
            & \textbf{\shortstack{Quality\\Score}}
            & \textbf{\shortstack{Semantic\\Score}}
            & \\ 
            \midrule 
            Wan2.1-T2V-1.3B & 91.83 & 94.71 & 99.17 & 96.51 & 65.00 & 76.09 & 89.93 & 74.60 & 53.66 & 20.03 & 62.37 & 79.81 & 65.43 & 76.93 \\
            Wan2.1-T2V-1.3B(FT.) & 93.59 & 95.81 & \textbf{99.36} & 97.21 & 54.08 & 79.90 & 88.57 & 78.98 & 58.20 & 28.55 & 63.31 & 81.26 & 68.47 & 78.71 \\
            VideoJAM(Reimpl.) & 91.51 & \textbf{96.01} & 99.13 & 96.05 & 73.88 & 79.22 & 88.92 & 79.20 & 59.66 & 28.34 & 66.17 & 81.18 & 69.08 & 78.76 \\
            \midrule
            \textbf{\small DreamWorld (Ours)} & \textbf{93.62} & 94.95 & 98.81 & \textbf{98.07} & \textbf{79.16} & \textbf{81.32} & \textbf{92.61} & \textbf{81.20} & \textbf{65.03} & \textbf{29.71} & \textbf{70.47} & \textbf{83.49} & \textbf{70.89} & \textbf{80.97} \\
            \bottomrule
        \end{tabular}%
    }
    % \vspace{-10pt}
\end{table*}

\begin{table*}[tb!]
    \centering
\begin{minipage}{0.48\linewidth}
    \centering
    \caption{\textbf{VBench-2.0 quantitative results.} \textbf{Control.}: Controllability; \textbf{Hum. Fid.}: Human Fidelity. Bold and \underline{underline} denote the best and \underline{second-best} results.}
    \label{tab:vbench2}
    \renewcommand{\arraystretch}{1.058} 
    \footnotesize 
    \setlength{\tabcolsep}{2.5pt}
    \resizebox{\linewidth}{!}{%
        \begin{tabular}{lccccc}
            \toprule
            \multirow{2}{*}[-2pt]{\textbf{\large Method}} & 
            \multicolumn{4}{c}{\textbf{\large VBench 2.0 Dimensions}} & 
            \multirow{2}{*}[-0.5ex]{\textbf{\large Overall}} \\
            \cmidrule(lr){2-5}
            & \textbf{ Creativity} 
            & \textbf{ Control.} 
            & \textbf{ Hum. Fid.} 
            & \textbf{ Physics} 
            & \\ 
            \midrule
            Wan2.1-T2V-1.3B      & 45.92 & 16.81 & 76.09 & \textbf{55.85} & 50.77 \\
            Wan2.1-T2V-1.3B(FT.)  & 43.13 & \textbf{18.41} & 77.09 & 54.51 & 51.18 \\
            VideoJAM             & \underline{49.32} & 16.33 & \underline{78.68} & 52.42 & \underline{52.33} \\
            \midrule
            \textbf{\large DreamWorld (Ours)} & \textbf{50.89} & \underline{16.95} & \textbf{80.11} & \underline{55.07} & \textbf{52.97} \\
            \bottomrule
        \end{tabular}%
    }
\end{minipage}\hfill
    \begin{minipage}{0.48\linewidth}
        \centering
        \caption{\textbf{Quantitative comparison on VideoPhy benchmark.} 
        We report Semantic Adherence (SA) and Physical Commonsense (PC) scores. 
        The best results are highlighted in \textbf{bold}.}
        % \vspace{0pt}
        \label{tab:videophy}
        \setlength{\tabcolsep}{3pt}
        \resizebox{\linewidth}{!}{%
            \begin{tabular}{lcccccccc}
                \toprule
                \multirow{2}{*}[-2pt]{\textbf{\large Method}} & 
                \multicolumn{2}{c}{\textbf{Solid-Solid}} & 
                \multicolumn{2}{c}{\textbf{Solid-Fluid}} & 
                \multicolumn{2}{c}{\textbf{Fluid-Fluid}} & 
                \multicolumn{2}{c}{\textbf{Overall}} \\
                \cmidrule(lr){2-3} \cmidrule(lr){4-5} \cmidrule(lr){6-7} \cmidrule(lr){8-9}
                & \textbf{SA} & \textbf{PC} & \textbf{SA} & \textbf{PC} & \textbf{SA} & \textbf{PC} & \textbf{SA} & \textbf{PC} \\
                \midrule
                Wan2.1-T2V-1.3B      & 51.1 & 22.3 & 45.2 & 19.1 & 45.4 & 23.6 & 47.7 & 21.2 \\
                Wan2.1-T2V-1.3B(FT.)  & 46.2 & 18.2 & 47.8 & 22.6 & 50.9 & 23.6 & 45.1 & 20.9 \\
                VideoJAM             & 51.7 & 29.4 & 38.4 & 23.3 & 47.3 & 20.1 & 45.3 & 25.3 \\
                \midrule
                \textbf{\large DreamWorld (Ours)} & \textbf{54.5} & \textbf{24.5} & \textbf{48.6} & \textbf{25.4} & \textbf{60.1} & \textbf{32.7} & \textbf{52.9} & \textbf{26.2} \\
                \bottomrule
            \end{tabular}%
        }
    \end{minipage}
\end{table*}

\section{Experimental Results}

\label{sec:experiments}

\subsection{Implementation Details}
\vspace{-4pt}
\label{subsec:implementation}

\paragraph{Model Configuration.}The proposed framework is established upon the pre-trained \textbf{Wan2.1-T2V-1.3B}~\cite{wan2025wanopenadvancedlargescale} architecture. The model is configured to synthesize video sequences comprising 81 frames with a spatial resolution of $480 \times 832$. To encapsulate multi-modal world knowledge priors, the joint feature stack employs the pre-trained RAFT model for optical flow estimation, while semantic and spatial representations are extracted via DINOv2~\cite{oquab2024dinov2learningrobustvisual} and VGGT~\cite{wang2025vggtvisualgeometrygrounded}, respectively.

\paragraph{Training Details.}For training protocol, the WISA~\cite{wang2025wisa} open-source dataset is utilized. All video samples are uniformly sampled for 81 frames, and then center-cropped and resized to align with the target resolution of $480 \times 832$. The model was efficiently fine-tuned using LoRA on a subset of 32k WISA videos, with a total of 2,000 optimization steps. The experiments are conducted on a cluster of 8 NVIDIA A100 GPUs with a total batch size of 16. For further elaboration on hyperparameters and architectural specifications, please refer to Appendix \ref{app:B}.

\vspace{-4pt}
\subsection{Quantitative Results}
\vspace{-4pt}
\label{subsec:quantitative}
% For a holistic assessment of generated video quality, the evaluation protocol incorporates the industry-standard VBench and its latest iteration VBench 2.0, VideoPhy, and the unified world generation benchmark, WorldScore.
For a comprehensive evaluation of generated video quality, we use the industry-standard VBench, its latest iteration VBench 2.0, VideoPhy, and the unified world generation benchmark, WorldScore. Additionally, all test prompts in our experiments are strictly based on the official prompt suites from the respective benchmark repositories.
\paragraph{VBench.}
The generative capability is systematically evaluated using VBench~\cite{huang2023vbenchcomprehensivebenchmarksuite}, a comprehensive benchmark assessing video quality and semantic consistency. As detailed in Table~\ref{tab:vbench}, DreamWorld achieves an overall score of 80.97, outperforming both the fine-tuned Wan2.1-T2V-1.3B baseline and VideoJAM, which aligns directly with our expectations for the injection of world knowledge. In the \textbf{temporal} dimension, kinematic priors significantly boost Motion Smoothness to 98.07 and Dynamic Degree to 79.16, indicating a better grasp of physical motion. Within the \textbf{semantic} and \textbf{spatial} domains, robust semantic and geometric constraints directly elevate the scores for Multiple Objects, Scene, and spatial Relationship to 65.03, 29.71, and 70.47, respectively. This confirms that incorporating explicit world knowledge successfully grounds the model's understanding of complex entities and spatial interactions, refining visual details and coherence rather than introducing artifacts. For a detailed breakdown across all 16 sub-dimensions, please refer to Appendix \ref{app:C}.
\paragraph{VBench 2.0.} 
% We extend our evaluation to VBench 2.0~\cite{zheng2025vbench20advancingvideogeneration}, which assesses human perceptual preferences across complex motion and compositional tasks. As shown in Table~\ref{tab:vbench2}, \textbf{DreamWorld} achieves a leading aggregate score of \textbf{52.97}, outperforming both the Wan2.1 baselines and VideoJAM. Our model excels in high-fidelity synthesis of diverse, semantically rich content without compromising dynamics, ultimately striking a superior balance between generative freedom and controllability.
Evaluations are extended to the rigorous VBench 2.0 framework~\cite{zheng2025vbench20advancingvideogeneration}, which employs a refined protocol designed to mirror human perceptual preferences across complex motion and compositional tasks. As quantitative results in Table~\ref{tab:vbench2} demonstrate, \textbf{DreamWorld} secures the leading position with an aggregate score of \textbf{53.64}, outperforming both the large-scale Wan2.1 baselines and the recent VideoJAM. The model proves particularly adept at high-fidelity synthesis highlighting its capability to generate diverse, semantically rich content. Crucially, this aesthetic superiority does not come at the expense of dynamics; our method maintains a highly competitive  score, significantly ahead of VideoJAM and Baseline, ultimately striking a superior balance between generative freedom and controllability.
\paragraph{VideoPhy.}
% We further evaluate on VideoPhy~\cite{bansal2024videophyevaluatingphysicalcommonsense}, a benchmark assessing physical commonsense in generated videos via Semantic Adherence (SA) and Physical Commonsense (PC) metrics. Following WISA~\cite{wang2025wisa}, we binarize SA and PC scores using a 0.5 threshold. As shown in Table~\ref{tab:videophy}, \textbf{DreamWorld} outperforms both the Baseline and the motion-prior-based VideoJAM. Specifically, our model achieves a state-of-the-art SA of 52.9\% and PC of 26.2\%, validating that our world-aware constraints effectively enhance the realism of dynamic interactions while preserving semantic fidelity.

We further extend our evaluation to VideoPhy~\cite{bansal2024videophyevaluatingphysicalcommonsense}, a rigorous benchmark designed to assess physical commonsense in generated videos. Unlike traditional metrics, VideoPhy specifically scrutinizes the model's ability to adhere to physical laws through Semantic Adherence (SA) and Physical Commonsense (PC), which evaluates compliance with real-world physics. Following WISA~\cite{wang2025wisa}, when the values of SA and PC are greater than or equal to 0.5, we set them as SA = 1 and PC =1. Values less than 0.5 are set as SA = 0 and PC = 0.

As presented in Table~\ref{tab:videophy}, \textbf{DreamWorld} outperforms both the standard Baseline and the motion-prior-based VideoJAM across aggregated metrics. Specifically, our model achieves a state-of-the-art  SA of 52.9\% and PC of 26.2\%, significantly surpassing other leading method. This improvement validates that our world-aware constraints not only enhance the realism of dynamic interactions but also preserve the semantic fidelity of the generation.
\begin{table}[t]
\centering
\begin{minipage}[t]{0.57\linewidth}
    \centering
    \caption{\textbf{Quantitative comparison on WorldScore}. Designed to assess the capabilities of world models, this benchmark evaluates simulation logic through static and dynamic scores. The best and \underline{second-best} results are highlighted in \textbf{bold} and \underline{underlined}.}
    \vspace{6pt}
    \label{tab:worldscore}
    \setlength{\tabcolsep}{3pt}
    \resizebox{\linewidth}{!}{%
    \renewcommand{\arraystretch}{1.43}
        \begin{tabular}{lcccccc}
            \toprule
            \multirow{2}{*}[-1.5ex]{\textbf{\large Method}} & 
            \multicolumn{3}{c}{\textbf{\large Static}} & 
            \multicolumn{2}{c}{\textbf{\large Dynamic}} &
            \multirow{2}{*}[-2ex]{\textbf{\shortstack{Overall\\Score}}} \\
            
            \cmidrule(lr){2-4}  \cmidrule(lr){5-6} 
            & \textbf{\scriptsize \shortstack{3D\\Consistency}} 
            & \textbf{\scriptsize \shortstack{Photometric\\Consistency}} 
            & \textbf{\scriptsize \shortstack{Subjective\\Quality}} 
            & \textbf{\scriptsize \shortstack{Motion\\Accuracy}} 
            & \textbf{\scriptsize \shortstack{Motion\\Smoothness}}
            &  \\ 
            \midrule
            Wan2.1-T2V-1.3B & \underline{72.31} & 74.41 & 52.21 & 32.36 & 68.18 & \underline{51.03} \\ 
            Wan2.1-T2V-1.3B(FT.) & 71.73 & 74.62 & \textbf{53.83} & \textbf{35.49} & 64.76 & 50.95 \\
            VideoJAM & 71.53 & \underline{75.39} & 44.03 & 33.38 & \textbf{71.95} & 49.38 \\
            \midrule
            \textbf{DreamWorld (Ours)} & \textbf{73.16} & \textbf{77.55} & \underline{52.45} & \underline{34.75} & \underline{69.02} & \textbf{51.48} \\
            \bottomrule
        \end{tabular}%
    }
\end{minipage}
\hfill
\begin{minipage}[t]{0.4\linewidth}
    \centering
     % We analyze the effectiveness of our method from two perspectives: (1) different \textbf{feature integration strategies}, and (2) the \textbf{Multi-Source Inner-Guidance} mechanism by selectively removing components.
    \caption{\textbf{Ablation studies.} We analyze our method's effectiveness through(1) different \textbf{feature integration strategies}, and (2)ablating the \textbf{Multi-Source Inner-Guidance} mechanism. }
    \vspace{6pt}
    \label{tab:ablation}
    \setlength{\tabcolsep}{4pt} 
    \resizebox{\linewidth}{!}{%
    \renewcommand{\arraystretch}{1.058}
    \begin{tabular}{lccc}
        \toprule
        \textbf{Method} & \textbf{Quality} & \textbf{Semantic} & \textbf{Overall} \\
        \midrule
        \small VGGT Only      & 81.76 & 71.36 & 79.68 \\
        \small VGGT + DINOv2  & 82.08 & 71.58 & 79.98 \\
        \midrule
        \small w/o Text Guidance   & 77.09 & 47.41 & 71.15 \\
        \small w/o VGGT Guidance   & 81.84 & 70.87 & 79.65 \\
        \small w/o DINOv2 Guidance & 81.98 & 70.95 & 79.78 \\
        \small w/o Optical Flow    & 81.87 & \textbf{71.06} & 79.71 \\
        \midrule
        \textbf{DreamWorld (Ours)} & \textbf{83.49} & 70.89 & \textbf{80.97} \\
        \bottomrule
    \end{tabular}
    }
\end{minipage}

\end{table}
\paragraph{WorldScore.}
To comprehensively assess holistic simulation capabilities, we benchmark DreamWorld using WorldScore~\cite{duan2025worldscoreunifiedevaluationbenchmark}, a unified benchmark tailored for the rigorous assessment of world simulators. As presented in Table~\ref{tab:worldscore}, our framework yields the highest overall score of 51.48. Beyond the aggregate score, the fine-grained evaluation explicitly demonstrates the substantial benefits of integrating world knowledge. By grounding the generation process in semantic and geometric priors, DreamWorld achieves remarkable static performance, highlighted by its leading \textit{3D Consistency} of 73.16 and \textit{Photometric Consistency} of 77.55, reflecting a profound understanding of structural and appearance preservation. Furthermore, rather than compromising temporal evolution for spatial quality, the embedded kinematic constraints ensure robust \textit{Motion Accuracy} and \textit{Motion Smoothness}. This seamlessly balanced performance across both static and dynamic axes underscores the efficacy of our multidimensional approach in simulating a realistic world.

\subsection{Qualitative Results}
\label{subsec:qualitative}
\begin{figure*}[tb!]
    \centering
    \includegraphics[width=\textwidth]{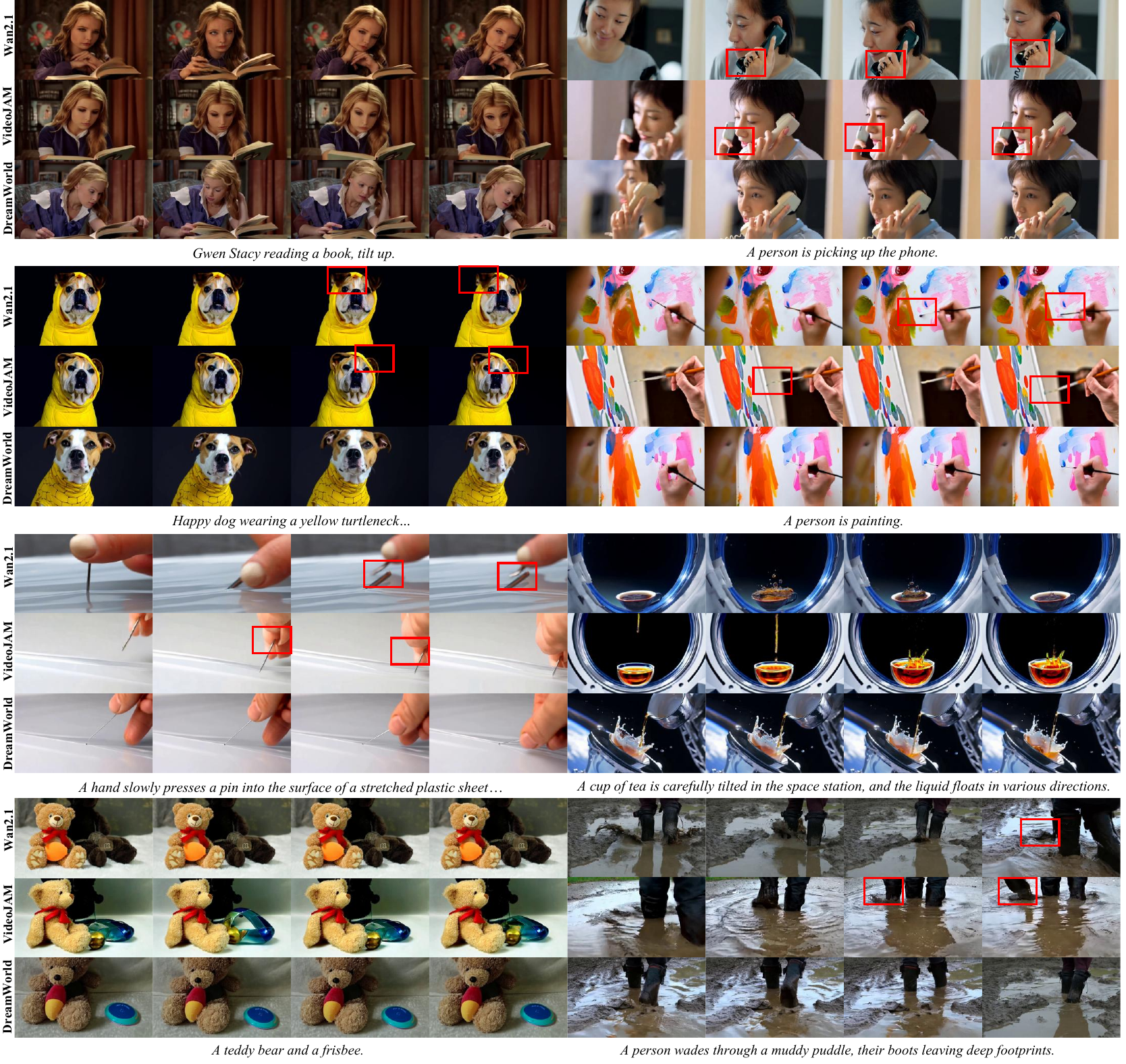}
    % \vspace{-20pt}
    \caption{\textbf{Qualitative comparison.} DreamWorld outperforms baselines by maintaining semantic realism, spatial integrity, and temporal precision. In contrast, competitor models frequently exhibit geometric penetrations and uncanny distortions}
    \label{fig:qualitative_comparison}
    % \vspace{-5pt}
\end{figure*}

% As illustrated in Figure~\ref{fig:qualitative_comparison}, we compare \textbf{DreamWorld} against VideoJAM and Wan2.1, demonstrating superior world consistency across three dimensions. Semantically, in the space station scenario, our model accurately tilts the cup for natural liquid flow, whereas baselines fail to initiate this critical action. Spatially, the dog example exhibits excellent 3D occlusion without physically impossible penetrations. Temporally, the Gwen Stacy example shows smooth motion and stable facial identity, avoiding common baseline deformations. These results demonstrate that injecting multi-modal world knowledge effectively constrains the diffusion process to strictly align with real-world logic (see Appendix~\ref{app:C} for further comparisons).
We compare DreamWorld against VideoJAM and Wan2.1 to evaluate world modeling capabilities. As illustrated in Figure~\ref{fig:qualitative_comparison}, our method exhibits superior world consistency. Semantically, in the space station scenario, our model accurately follows the prompt to carefully tilt the cup, allowing the liquid to flow down naturally; in contrast, both VideoJAM and Baseline fail to initiate the critical tilting action. Spatially, the happy dog demonstrates excellent 3D occlusion, ensuring the ears and turtleneck sweater are correctly positioned without physically impossible penetration. Temporally, as seen in the Gwen Stacy reading example, our model generates smooth and natural motion consistent and the character's facial identity remains stable, avoiding the temporal deformations often observed in baseline methods. These results demonstrate that injecting multi-modal world knowledge effectively constrains the diffusion process, ensuring strict alignment with real-world logic. Please refer to Appendix \ref{app:D} for further qualitative comparisons.
\subsection{Ablation Studies}
\label{subsec:ablation}
% \begin{figure}[t]
%     \centering
%     \includegraphics[width=1.0\linewidth]{fig/lambda_ablation.pdf}
%     % \vspace{-15pt}
%     \caption{\textbf{Influence of loss weights $\lambda$.} Quantitative comparison of generation quality and semantic alignment across different weight settings. The results indicate that $\lambda=0.2$ yields the best trade-off.}
%     % \vspace{-15pt}
%     \label{fig:λ_metrics}
% \end{figure}

The efficacy of DreamWorld is validated through a three-dimensional analysis, with VBench consistently employed as the default benchmark for ablation experiments, barring any specific exceptions.
\paragraph{Effectiveness of World Knowledge} 
The necessity of the comprehensive WorldKnowledge stack is rigorously validated by progressively evaluating its constituent feature subsets. As shown in Table~\ref{tab:ablation}, relying on the spatial prior (\textbf{VGGT Only}) provides basic geometric grounding but lacks semantic coherence. The integration of the semantic prior(\textbf{VGGT + DINOv2}) significantly bolsters the model's capabilities, particularly in maintaining object permanence and ensuring precise text-video alignment. Nevertheless, the optimal performance is achieved only by the full model, confirming that the synergy of spatial guidance, semantic context, and temporal consistency is essential for robust and realistic world simulation.

\begin{figure}[tb!]
    \centering
    \begin{minipage}{0.49\linewidth}
        \centering
        \includegraphics[width=\linewidth]{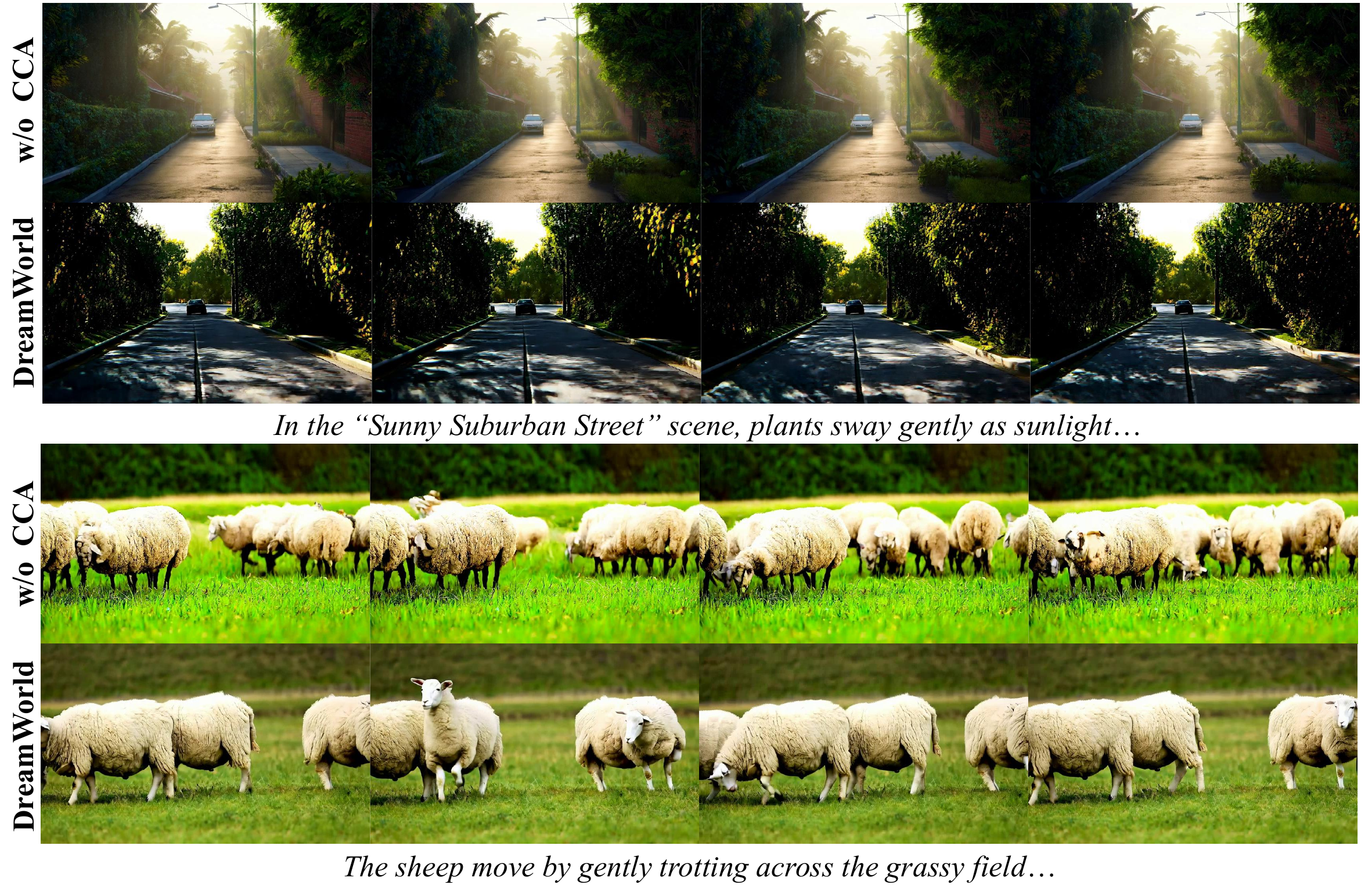}
        \caption{\textbf{Effectiveness of CCA.} Without CCA, the generated videos suffer from severe flickering and abnormal highlighting artifacts, leading to degraded visual quality.}
        \label{fig:cca_ablation}
    \end{minipage}
    \hfill
    \begin{minipage}{0.49\linewidth}
        \centering
        \includegraphics[width=\linewidth]{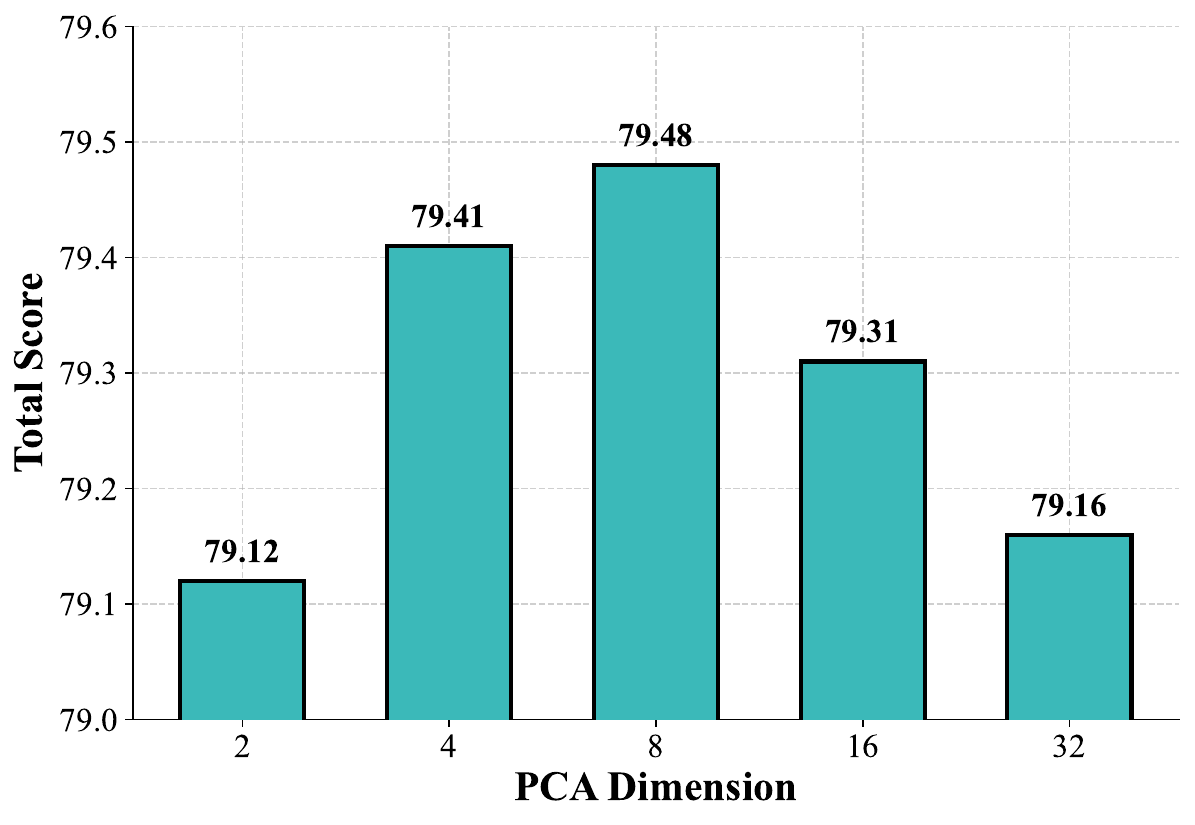}
        \vspace{-18pt}
        \caption{\textbf{Influence of Dimensionality reduction.} Ablations on the PCA dimensionality using a 10k dataset demonstrate that a reduction to 8 dimensions yields the optimal performance.}
        \label{fig:pca_ablation}
    \end{minipage}
\end{figure}

\paragraph{Effectiveness of CCA.} 
The qualitative efficacy of Consistent Constraint Annealing (CCA) is evidenced by the visual artifacts observed in the static weighting baseline, as shown in Figure~\ref{fig:cca_ablation}. Without the dynamic schedule, optimization conflicts manifest as severe abnormal highlighting and exposure anomalies, particularly visible in the  Balcony scenes where lighting becomes unnaturally intense. Furthermore, temporal instability is evident in the Moving Sheep, which suffers from high-frequency flickering and texture jittering. The proposed CCA strategy effectively eliminates these artifacts, allowing the model to refine visual details without interference, thus ensuring both photorealism and temporal smoothness.More ablation experiments are provided in Appendix \ref{app:E}.
\paragraph{Effectiveness of Multi-Source Inner-Guidance.}
We investigate the necessity of integrating diverse priors through an ablation study on the Multi-Source Inner-Guidance mechanism (Table \ref{tab:ablation}). The results indicate that removing any individual component leads to consistent performance degradation. Notably, excluding Temporal Priors causes the most significant drop in Quality and Overall scores, the absence of Text Guidance or Semantic Priors severely impairs the Semantic score. Ultimately,  DreamWorld achieves superior performance, confirming that the synergistic integration of text, semantic, spatial, and temporal cues is essential for a robust physics-aware world model.
% \paragraph{Influence of Dimensionality reduction.}
% We investigate the sensitivity of our model to the  loss weight $\lambda$, which controls the strength of the world knowledge priors. To quantify the trade-off between generative fidelity and condition adherence, we report both the Quality Score and Semantic Score on VBench across varying $\lambda$ values. As illustrated in Fig.~\ref{fig:lambda_metrics}, the performance of both metrics peaks at $\lambda=0.2$. We observe that lower values fail to effectively enforce the physical and semantic constraints, whereas excessive weights interfere with the diffusion backbone's distribution, leading to a degradation in visual quality. Consequently, we adopt $\lambda=0.2$ as the optimal configuration for our full model.
% \paragraph{Influence of Dimensionality reduction.} We investigate the sensitivity of our model to the PCA dimensionality, which controls the information bottleneck of world knowledge priors. To quantify this impact, we report the VBench Total Score across varying dimensions. As illustrated in Fig.~\ref{fig:pca_ablation}, performance peaks at a dimensionality of 8 with a score of 79.48. We observe that lower dimensions result in over-compression and a loss of critical details, whereas excessively high dimensions retain redundant noise that interferes with the diffusion backbone, leading to a degradation in visual quality. Consequently, we adopt a PCA dimension of 8 as the optimal configuration for our full model.
\paragraph{Influence of Dimensionality reduction.}We investigate the sensitivity of our model to the PCA dimensionality, which controls the information bottleneck of world knowledge priors. To quantify this impact, we report the VBench Total Score across varying dimensions.As illustrated in Figure~\ref{fig:pca_ablation}, the performance peaks at a dimensionality of 8, achieving a Total Score of 79.48. We observe that lower dimensions (e.g., 2 or 4) result in over-compression, failing to retain critical semantic and geometric details necessary for effective structural guidance. Conversely, excessively high dimensions retain redundant information, which complicates the joint feature integration within the diffusion backbone and leads to a degradation in overall synthesis quality. Consequently, we adopt a PCA dimension of 8 as the optimal configuration for our expert feature representations.

\section{Conclusion}
\label{sec:conclu}
In this work, we presented \textbf{DreamWorld}, a unified framework designed to bridge the prevailing gap between high-fidelity video synthesis and intrinsic world coherence. We identify that previous approaches often focus on distilling a single expert model. However, directly enforcing simultaneous alignment across multiple knowledge sources is prone to collapsing into trivial solutions due to conflicting optimization objectives. We introduced a softer \textbf{Joint World Modeling Paradigm} that facilitates the synergistic integration of multi-source expert knowledge. This gentle alignment approach, complemented by our \textit{Consistent Constraint Annealing (CCA)} strategy and \textit{Multi-Source Inner-Guidance}, effectively harmonizes the intricate interplay between structural logic and generative freedom. DreamWorld establishes a new state-of-the-art on standard benchmarks, validating its potential as a robust foundation for next-generation general-purpose world simulators.

\textbf{Limitations.} Despite these advancements, our approach is currently constrained by the computational resources and the diversity of the training datasets. Future research could explore optimizing the efficiency of this multi-source integration and incorporating more diverse data curation to further enhance the universality of the simulated world.

\bibliography{neurips_2026}

@misc{oquab2024dinov2learningrobustvisual,
      title={DINOv2: Learning Robust Visual Features without Supervision}, 
      author={Maxime Oquab and Timothée Darcet and Théo Moutakanni and Huy Vo and Marc Szafraniec and Vasil Khalidov and Pierre Fernandez and Daniel Haziza and Francisco Massa and Alaaeldin El-Nouby and Mahmoud Assran and Nicolas Ballas and Wojciech Galuba and Russell Howes and Po-Yao Huang and Shang-Wen Li and Ishan Misra and Michael Rabbat and Vasu Sharma and Gabriel Synnaeve and Hu Xu and Hervé Jegou and Julien Mairal and Patrick Labatut and Armand Joulin and Piotr Bojanowski},
      year={2024},
      eprint={2304.07193},
      archivePrefix={arXiv},
      primaryClass={cs.CV},
      url={https://arxiv.org/abs/2304.07193}, 
}

@misc{wang2025vggtvisualgeometrygrounded,
      title={VGGT: Visual Geometry Grounded Transformer}, 
      author={Jianyuan Wang and Minghao Chen and Nikita Karaev and Andrea Vedaldi and Christian Rupprecht and David Novotny},
      year={2025},
      eprint={2503.11651},
      archivePrefix={arXiv},
      primaryClass={cs.CV},
      url={https://arxiv.org/abs/2503.11651}, 
}

@misc{liu2025alignvidtrainingfreeattentionscaling,
      title={AlignVid: Training-Free Attention Scaling for Semantic Fidelity in Text-Guided Image-to-Video Generation}, 
      author={Yexin Liu and Wen-Jie Shu and Zile Huang and Haoze Zheng and Yueze Wang and Manyuan Zhang and Ser-Nam Lim and Harry Yang},
      year={2025},
      eprint={2512.01334},
      archivePrefix={arXiv},
      primaryClass={cs.CV},
      url={https://arxiv.org/abs/2512.01334}, 
}

@misc{chefer2025videojamjointappearancemotionrepresentations,
      title={VideoJAM: Joint Appearance-Motion Representations for Enhanced Motion Generation in Video Models}, 
      author={Hila Chefer and Uriel Singer and Amit Zohar and Yuval Kirstain and Adam Polyak and Yaniv Taigman and Lior Wolf and Shelly Sheynin},
      year={2025},
      eprint={2502.02492},
      archivePrefix={arXiv},
      primaryClass={cs.CV},
      url={https://arxiv.org/abs/2502.02492}, 
}

@misc{teed2020raftrecurrentallpairsfield,
      title={RAFT: Recurrent All-Pairs Field Transforms for Optical Flow}, 
      author={Zachary Teed and Jia Deng},
      year={2020},
      eprint={2003.12039},
      archivePrefix={arXiv},
      primaryClass={cs.CV},
      url={https://arxiv.org/abs/2003.12039}, 
}

@misc{lipman2023flowmatchinggenerativemodeling,
      title={Flow Matching for Generative Modeling}, 
      author={Yaron Lipman and Ricky T. Q. Chen and Heli Ben-Hamu and Maximilian Nickel and Matt Le},
      year={2023},
      eprint={2210.02747},
      archivePrefix={arXiv},
      primaryClass={cs.LG},
      url={https://arxiv.org/abs/2210.02747}, 
}

@misc{bhowmik2025moalignmotioncentricrepresentationalignment,
      title={MoAlign: Motion-Centric Representation Alignment for Video Diffusion Models}, 
      author={Aritra Bhowmik and Denis Korzhenkov and Cees G. M. Snoek and Amirhossein Habibian and Mohsen Ghafoorian},
      year={2025},
      eprint={2510.19022},
      archivePrefix={arXiv},
      primaryClass={cs.CV},
      url={https://arxiv.org/abs/2510.19022}, 
}

@misc{min2024driveworld4dpretrainedscene,
      title={DriveWorld: 4D Pre-trained Scene Understanding via World Models for Autonomous Driving}, 
      author={Chen Min and Dawei Zhao and Liang Xiao and Jian Zhao and Xinli Xu and Zheng Zhu and Lei Jin and Jianshu Li and Yulan Guo and Junliang Xing and Liping Jing and Yiming Nie and Bin Dai},
      year={2024},
      eprint={2405.04390},
      archivePrefix={arXiv},
      primaryClass={cs.CV},
      url={https://arxiv.org/abs/2405.04390}, 
}

@misc{wan2025wanopenadvancedlargescale,
      title={Wan: Open and Advanced Large-Scale Video Generative Models}, 
      author={Team Wan and Ang Wang and Baole Ai and Bin Wen and Chaojie Mao and Chen-Wei Xie and Di Chen and Feiwu Yu and Haiming Zhao and Jianxiao Yang and Jianyuan Zeng and Jiayu Wang and Jingfeng Zhang and Jingren Zhou and Jinkai Wang and Jixuan Chen and Kai Zhu and Kang Zhao and Keyu Yan and Lianghua Huang and Mengyang Feng and Ningyi Zhang and Pandeng Li and Pingyu Wu and Ruihang Chu and Ruili Feng and Shiwei Zhang and Siyang Sun and Tao Fang and Tianxing Wang and Tianyi Gui and Tingyu Weng and Tong Shen and Wei Lin and Wei Wang and Wei Wang and Wenmeng Zhou and Wente Wang and Wenting Shen and Wenyuan Yu and Xianzhong Shi and Xiaoming Huang and Xin Xu and Yan Kou and Yangyu Lv and Yifei Li and Yijing Liu and Yiming Wang and Yingya Zhang and Yitong Huang and Yong Li and You Wu and Yu Liu and Yulin Pan and Yun Zheng and Yuntao Hong and Yupeng Shi and Yutong Feng and Zeyinzi Jiang and Zhen Han and Zhi-Fan Wu and Ziyu Liu},
      year={2025},
      eprint={2503.20314},
      archivePrefix={arXiv},
      primaryClass={cs.CV},
      url={https://arxiv.org/abs/2503.20314}, 
}

@misc{zheng2025diffusiontransformersrepresentationautoencoders,

      title={Diffusion Transformers with Representation Autoencoders}, 

      author={Boyang Zheng and Nanye Ma and Shengbang Tong and Saining Xie},

      year={2025},

      eprint={2510.11690},

      archivePrefix={arXiv},

      primaryClass={cs.CV},

      url={https://arxiv.org/abs/2510.11690}, 

}

@misc{jiang2025representationcomponentneededdiffusion,

      title={No Other Representation Component Is Needed: Diffusion Transformers Can Provide Representation Guidance by Themselves}, 

      author={Dengyang Jiang and Mengmeng Wang and Liuzhuozheng Li and Lei Zhang and Haoyu Wang and Wei Wei and Guang Dai and Yanning Zhang and Jingdong Wang},

      year={2025},

      eprint={2505.02831},

      archivePrefix={arXiv},

      primaryClass={cs.CV},

      url={https://arxiv.org/abs/2505.02831}, 

}

@misc{lee2025aligningtextimagediffusion,

      title={Aligning Text to Image in Diffusion Models is Easier Than You Think}, 

      author={Jaa-Yeon Lee and Byunghee Cha and Jeongsol Kim and Jong Chul Ye},

      year={2025},

      eprint={2503.08250},

      archivePrefix={arXiv},

      primaryClass={cs.CV},

      url={https://arxiv.org/abs/2503.08250}, 

}

@misc{li2025semanticspaceinterveneddiffusivealignmentvisual,

      title={Semantic-Space-Intervened Diffusive Alignment for Visual Classification}, 

      author={Zixuan Li and Lei Meng and Guoqing Chao and Wei Wu and Xiaoshuo Yan and Yimeng Yang and Zhuang Qi and Xiangxu Meng},

      year={2025},

      eprint={2505.05721},

      archivePrefix={arXiv},

      primaryClass={cs.CV},

      url={https://arxiv.org/abs/2505.05721}, 

}

@misc{chen2024diffusionforcingnexttokenprediction,

      title={Diffusion Forcing: Next-token Prediction Meets Full-Sequence Diffusion}, 

      author={Boyuan Chen and Diego Marti Monso and Yilun Du and Max Simchowitz and Russ Tedrake and Vincent Sitzmann},

      year={2024},

      eprint={2407.01392},

      archivePrefix={arXiv},

      primaryClass={cs.LG},

      url={https://arxiv.org/abs/2407.01392}, 

}

@misc{song2025historyguidedvideodiffusion,

      title={History-Guided Video Diffusion}, 

      author={Kiwhan Song and Boyuan Chen and Max Simchowitz and Yilun Du and Russ Tedrake and Vincent Sitzmann},

      year={2025},

      eprint={2502.06764},

      archivePrefix={arXiv},

      primaryClass={cs.LG},

      url={https://arxiv.org/abs/2502.06764}, 

}

@misc{fuest2025maskflowdiscreteflowsflexible,

      title={MaskFlow: Discrete Flows For Flexible and Efficient Long Video Generation}, 

      author={Michael Fuest and Vincent Tao Hu and Björn Ommer},

      year={2025},

      eprint={2502.11234},

      archivePrefix={arXiv},

      primaryClass={cs.CV},

      url={https://arxiv.org/abs/2502.11234}, 

}

@misc{wu2025videoworldmodelslongterm,

      title={Video World Models with Long-term Spatial Memory}, 

      author={Tong Wu and Shuai Yang and Ryan Po and Yinghao Xu and Ziwei Liu and Dahua Lin and Gordon Wetzstein},

      year={2025},

      eprint={2506.05284},

      archivePrefix={arXiv},

      primaryClass={cs.CV},

      url={https://arxiv.org/abs/2506.05284}, 

}

@misc{po2025longcontextstatespacevideoworld,

      title={Long-Context State-Space Video World Models}, 

      author={Ryan Po and Yotam Nitzan and Richard Zhang and Berlin Chen and Tri Dao and Eli Shechtman and Gordon Wetzstein and Xun Huang},

      year={2025},

      eprint={2505.20171},

      archivePrefix={arXiv},

      primaryClass={cs.CV},

      url={https://arxiv.org/abs/2505.20171}, 

}

@misc{xiao2026worldmemlongtermconsistentworld,

      title={WorldMem: Long-term Consistent World Simulation with Memory}, 

      author={Zeqi Xiao and Yushi Lan and Yifan Zhou and Wenqi Ouyang and Shuai Yang and Yanhong Zeng and Xingang Pan},

      year={2026},

      eprint={2504.12369},

      archivePrefix={arXiv},

      primaryClass={cs.CV},

      url={https://arxiv.org/abs/2504.12369}, 

}

@misc{zhao2025guidingcrossmodalrepresentationsmllm,

      title={Guiding Cross-Modal Representations with MLLM Priors via Preference Alignment}, 

      author={Pengfei Zhao and Rongbo Luan and Wei Zhang and Peng Wu and Sifeng He},

      year={2025},

      eprint={2506.06970},

      archivePrefix={arXiv},

      primaryClass={cs.CV},

      url={https://arxiv.org/abs/2506.06970}, 

}

@misc{berg2025semanticworldmodels,

      title={Semantic World Models}, 

      author={Jacob Berg and Chuning Zhu and Yanda Bao and Ishan Durugkar and Abhishek Gupta},

      year={2025},

      eprint={2510.19818},

      archivePrefix={arXiv},

      primaryClass={cs.LG},

      url={https://arxiv.org/abs/2510.19818}, 

}

@misc{fu2019cyclicalannealingschedulesimple,
      title={Cyclical Annealing Schedule: A Simple Approach to Mitigating KL Vanishing}, 
      author={Hao Fu and Chunyuan Li and Xiaodong Liu and Jianfeng Gao and Asli Celikyilmaz and Lawrence Carin},
      year={2019},
      eprint={1903.10145},
      archivePrefix={arXiv},
      primaryClass={cs.LG},
      url={https://arxiv.org/abs/1903.10145}, 
}

@misc{jang2025multidimensionalpreferencealignmentconditioning,
      title={Multi-dimensional Preference Alignment by Conditioning Reward Itself}, 
      author={Jiho Jang and Jinyoung Kim and Kyungjune Baek and Nojun Kwak},
      year={2025},
      eprint={2512.10237},
      archivePrefix={arXiv},
      primaryClass={cs.CV},
      url={https://arxiv.org/abs/2512.10237}, 
}

@misc{kouzelis2026boostinggenerativeimagemodeling,
      title={Boosting Generative Image Modeling via Joint Image-Feature Synthesis}, 
      author={Theodoros Kouzelis and Efstathios Karypidis and Ioannis Kakogeorgiou and Spyros Gidaris and Nikos Komodakis},
      year={2026},
      eprint={2504.16064},
      archivePrefix={arXiv},
      primaryClass={cs.CV},
      url={https://arxiv.org/abs/2504.16064}, 
}

@article{vincent2011connection,

  title={A connection between score matching and denoising autoencoders},

  author={Vincent, Pascal},

  journal={Neural computation},

  volume={23},

  number={7},

  pages={1661--1674},

  year={2011},

  publisher={MIT Press}

}

@misc{ho2022classifierfreediffusionguidance,

      title={Classifier-Free Diffusion Guidance}, 

      author={Jonathan Ho and Tim Salimans},

      year={2022},

      eprint={2207.12598},

      archivePrefix={arXiv},

      primaryClass={cs.LG},

      url={https://arxiv.org/abs/2207.12598}, 

}

@misc{wang2025wisa,

                title={WISA: World Simulator Assistant for Physics-Aware Text-to-Video Generation}, 

                author={Jing Wang and Ao Ma and Ke Cao and Jun Zheng and Zhanjie Zhang and Jiasong Feng and Shanyuan Liu and Yuhang Ma and Bo Cheng and Dawei Leng and Yuhui Yin and Xiaodan Liang},

                year={2025},

                eprint={2502.08153},

                archivePrefix={arXiv},

                primaryClass={cs.CV},

                url={https://arxiv.org/abs/2502.08153}, 

}

@misc{kang2025farvideogenerationworld,

      title={How Far is Video Generation from World Model: A Physical Law Perspective}, 

      author={Bingyi Kang and Yang Yue and Rui Lu and Zhijie Lin and Yang Zhao and Kaixin Wang and Gao Huang and Jiashi Feng},

      year={2025},

      eprint={2411.02385},

      archivePrefix={arXiv},

      primaryClass={cs.CV},

      url={https://arxiv.org/abs/2411.02385}, 

}

@misc{yu2025representationalignmentgenerationtraining,
      title={Representation Alignment for Generation: Training Diffusion Transformers Is Easier Than You Think}, 
      author={Sihyun Yu and Sangkyung Kwak and Huiwon Jang and Jongheon Jeong and Jonathan Huang and Jinwoo Shin and Saining Xie},
      year={2025},
      eprint={2410.06940},
      archivePrefix={arXiv},
      primaryClass={cs.CV},
      url={https://arxiv.org/abs/2410.06940}, 
}

@inproceedings{yu2025repa,
  title={Representation Alignment for Generation: Training Diffusion Transformers Is Easier Than You Think},
  author={Sihyun Yu and Sangkyung Kwak and Huiwon Jang and Jongheon Jeong and Jonathan Huang and Jinwoo Shin and Saining Xie},
  year={2025},
  booktitle={International Conference on Learning Representations},
}

@misc{zhang2025videorepalearningphysicsvideo,
      title={VideoREPA: Learning Physics for Video Generation through Relational Alignment with Foundation Models}, 
      author={Xiangdong Zhang and Jiaqi Liao and Shaofeng Zhang and Fanqing Meng and Xiangpeng Wan and Junchi Yan and Yu Cheng},
      year={2025},
      eprint={2505.23656},
      archivePrefix={arXiv},
      primaryClass={cs.CV},
      url={https://arxiv.org/abs/2505.23656}, 
}

@misc{duan2025worldscoreunifiedevaluationbenchmark,
      title={WorldScore: A Unified Evaluation Benchmark for World Generation}, 
      author={Haoyi Duan and Hong-Xing Yu and Sirui Chen and Li Fei-Fei and Jiajun Wu},
      year={2025},
      eprint={2504.00983},
      archivePrefix={arXiv},
      primaryClass={cs.GR},
      url={https://arxiv.org/abs/2504.00983}, 
}

@misc{huang2023vbenchcomprehensivebenchmarksuite,
      title={VBench: Comprehensive Benchmark Suite for Video Generative Models}, 
      author={Ziqi Huang and Yinan He and Jiashuo Yu and Fan Zhang and Chenyang Si and Yuming Jiang and Yuanhan Zhang and Tianxing Wu and Qingyang Jin and Nattapol Chanpaisit and Yaohui Wang and Xinyuan Chen and Limin Wang and Dahua Lin and Yu Qiao and Ziwei Liu},
      year={2023},
      eprint={2311.17982},
      archivePrefix={arXiv},
      primaryClass={cs.CV},
      url={https://arxiv.org/abs/2311.17982}, 
}

@misc{zheng2025vbench20advancingvideogeneration,
      title={VBench-2.0: Advancing Video Generation Benchmark Suite for Intrinsic Faithfulness}, 
      author={Dian Zheng and Ziqi Huang and Hongbo Liu and Kai Zou and Yinan He and Fan Zhang and Lulu Gu and Yuanhan Zhang and Jingwen He and Wei-Shi Zheng and Yu Qiao and Ziwei Liu},
      year={2025},
      eprint={2503.21755},
      archivePrefix={arXiv},
      primaryClass={cs.CV},
      url={https://arxiv.org/abs/2503.21755}, 
}

@article{bardes2024revisiting,
  title={Revisiting Feature Prediction for Learning Visual Representations from Video},
  author={Bardes, Adrien and Garrido, Quentin and Ponce, Jean and Rabbat, Michael and LeCun, Yann and Assran, Mahmoud and Ballas, Nicolas},
  journal={arXiv:2404.08471},
  year={2024}
}

@misc{bansal2024videophyevaluatingphysicalcommonsense,
      title={VideoPhy: Evaluating Physical Commonsense for Video Generation}, 
      author={Hritik Bansal and Zongyu Lin and Tianyi Xie and Zeshun Zong and Michal Yarom and Yonatan Bitton and Chenfanfu Jiang and Yizhou Sun and Kai-Wei Chang and Aditya Grover},
      year={2024},
      eprint={2406.03520},
      archivePrefix={arXiv},
      primaryClass={cs.CV},
      url={https://arxiv.org/abs/2406.03520}, 
}

@article{videoworldsimulators2024,
  title={Video generation models as world simulators},
  author={Tim Brooks and Bill Peebles and Connor Holmes and Will DePue and Yufei Guo and Li Jing and David Schnurr and Joe Taylor and Troy Luhman and Eric Luhman and Clarence Ng and Ricky Wang and Aditya Ramesh},
  year={2024},
  url={https://openai.com/research/video-generation-models-as-world-simulators},
}

@misc{qin2024worldsimbenchvideogenerationmodels,
      title={WorldSimBench: Towards Video Generation Models as World Simulators}, 
      author={Yiran Qin and Zhelun Shi and Jiwen Yu and Xijun Wang and Enshen Zhou and Lijun Li and Zhenfei Yin and Xihui Liu and Lu Sheng and Jing Shao and Lei Bai and Wanli Ouyang and Ruimao Zhang},
      year={2024},
      eprint={2410.18072},
      archivePrefix={arXiv},
      primaryClass={cs.CV},
      url={https://arxiv.org/abs/2410.18072}, 
}

@article{HORN1981185,
title = {Determining optical flow},
journal = {Artificial Intelligence},
volume = {17},
number = {1},
pages = {185-203},
year = {1981},
issn = {0004-3702},
doi = {https://doi.org/10.1016/0004-3702(81)90024-2},
url = {https://www.sciencedirect.com/science/article/pii/0004370281900242},
author = {Berthold K.P. Horn and Brian G. Schunck}
}

@misc{peebles2023scalablediffusionmodelstransformers,
      title={Scalable Diffusion Models with Transformers}, 
      author={William Peebles and Saining Xie},
      year={2023},
      eprint={2212.09748},
      archivePrefix={arXiv},
      primaryClass={cs.CV},
      url={https://arxiv.org/abs/2212.09748}, 
}

@misc{tong2025thinkingvideovideogeneration,
      title={Thinking with Video: Video Generation as a Promising Multimodal Reasoning Paradigm}, 
      author={Jingqi Tong and Yurong Mou and Hangcheng Li and Mingzhe Li and Yongzhuo Yang and Ming Zhang and Qiguang Chen and Tianyi Liang and Xiaomeng Hu and Yining Zheng and Xinchi Chen and Jun Zhao and Xuanjing Huang and Xipeng Qiu},
      year={2025},
      eprint={2511.04570},
      archivePrefix={arXiv},
      primaryClass={cs.CV},
      url={https://arxiv.org/abs/2511.04570}, 
}

@misc{kaplan2020scalinglawsneurallanguage,
      title={Scaling Laws for Neural Language Models}, 
      author={Jared Kaplan and Sam McCandlish and Tom Henighan and Tom B. Brown and Benjamin Chess and Rewon Child and Scott Gray and Alec Radford and Jeffrey Wu and Dario Amodei},
      year={2020},
      eprint={2001.08361},
      archivePrefix={arXiv},
      primaryClass={cs.LG},
      url={https://arxiv.org/abs/2001.08361}, 
}

@misc{ma2025lattelatentdiffusiontransformer,
      title={Latte: Latent Diffusion Transformer for Video Generation}, 
      author={Xin Ma and Yaohui Wang and Xinyuan Chen and Gengyun Jia and Ziwei Liu and Yuan-Fang Li and Cunjian Chen and Yu Qiao},
      year={2025},
      eprint={2401.03048},
      archivePrefix={arXiv},
      primaryClass={cs.CV},
      url={https://arxiv.org/abs/2401.03048}, 
}

@misc{bartal2024lumierespacetimediffusionmodel,
      title={Lumiere: A Space-Time Diffusion Model for Video Generation}, 
      author={Omer Bar-Tal and Hila Chefer and Omer Tov and Charles Herrmann and Roni Paiss and Shiran Zada and Ariel Ephrat and Junhwa Hur and Guanghui Liu and Amit Raj and Yuanzhen Li and Michael Rubinstein and Tomer Michaeli and Oliver Wang and Deqing Sun and Tali Dekel and Inbar Mosseri},
      year={2024},
      eprint={2401.12945},
      archivePrefix={arXiv},
      primaryClass={cs.CV},
      url={https://arxiv.org/abs/2401.12945}, 
}

@misc{zheng2024opensorademocratizingefficientvideo,
      title={Open-Sora: Democratizing Efficient Video Production for All}, 
      author={Zangwei Zheng and Xiangyu Peng and Tianji Yang and Chenhui Shen and Shenggui Li and Hongxin Liu and Yukun Zhou and Tianyi Li and Yang You},
      year={2024},
      eprint={2412.20404},
      archivePrefix={arXiv},
      primaryClass={cs.CV},
      url={https://arxiv.org/abs/2412.20404}, 
}

@misc{kong2025hunyuanvideosystematicframeworklarge,

      title={HunyuanVideo: A Systematic Framework For Large Video Generative Models}, 

      author={Weijie Kong and Qi Tian and Zijian Zhang and Rox Min and Zuozhuo Dai and Jin Zhou and Jiangfeng Xiong and Xin Li and Bo Wu and Jianwei Zhang and Kathrina Wu and Qin Lin and Junkun Yuan and Yanxin Long and Aladdin Wang and Andong Wang and Changlin Li and Duojun Huang and Fang Yang and Hao Tan and Hongmei Wang and Jacob Song and Jiawang Bai and Jianbing Wu and Jinbao Xue and Joey Wang and Kai Wang and Mengyang Liu and Pengyu Li and Shuai Li and Weiyan Wang and Wenqing Yu and Xinchi Deng and Yang Li and Yi Chen and Yutao Cui and Yuanbo Peng and Zhentao Yu and Zhiyu He and Zhiyong Xu and Zixiang Zhou and Zunnan Xu and Yangyu Tao and Qinglin Lu and Songtao Liu and Dax Zhou and Hongfa Wang and Yong Yang and Di Wang and Yuhong Liu and Jie Jiang and Caesar Zhong},

      year={2025},

      eprint={2412.03603},

      archivePrefix={arXiv},

      primaryClass={cs.CV},

      url={https://arxiv.org/abs/2412.03603}, 

}

@misc{yang2025cogvideoxtexttovideodiffusionmodels,

      title={CogVideoX: Text-to-Video Diffusion Models with An Expert Transformer}, 

      author={Zhuoyi Yang and Jiayan Teng and Wendi Zheng and Ming Ding and Shiyu Huang and Jiazheng Xu and Yuanming Yang and Wenyi Hong and Xiaohan Zhang and Guanyu Feng and Da Yin and Yuxuan Zhang and Weihan Wang and Yean Cheng and Bin Xu and Xiaotao Gu and Yuxiao Dong and Jie Tang},

      year={2025},

      eprint={2408.06072},

      archivePrefix={arXiv},

      primaryClass={cs.CV},

      url={https://arxiv.org/abs/2408.06072}, 

}

@misc{hacohen2026ltx2efficientjointaudiovisual,

      title={LTX-2: Efficient Joint Audio-Visual Foundation Model}, 

      author={Yoav HaCohen and Benny Brazowski and Nisan Chiprut and Yaki Bitterman and Andrew Kvochko and Avishai Berkowitz and Daniel Shalem and Daphna Lifschitz and Dudu Moshe and Eitan Porat and Eitan Richardson and Guy Shiran and Itay Chachy and Jonathan Chetboun and Michael Finkelson and Michael Kupchick and Nir Zabari and Nitzan Guetta and Noa Kotler and Ofir Bibi and Ori Gordon and Poriya Panet and Roi Benita and Shahar Armon and Victor Kulikov and Yaron Inger and Yonatan Shiftan and Zeev Melumian and Zeev Farbman},

      year={2026},

      eprint={2601.03233},

      archivePrefix={arXiv},

      primaryClass={cs.CV},

      url={https://arxiv.org/abs/2601.03233}, 

}

@misc{jin2025pyramidalflowmatchingefficient,

      title={Pyramidal Flow Matching for Efficient Video Generative Modeling}, 

      author={Yang Jin and Zhicheng Sun and Ningyuan Li and Kun Xu and Kun Xu and Hao Jiang and Nan Zhuang and Quzhe Huang and Yang Song and Yadong Mu and Zhouchen Lin},

      year={2025},

      eprint={2410.05954},

      archivePrefix={arXiv},

      primaryClass={cs.CV},

      url={https://arxiv.org/abs/2410.05954}, 

}

@misc{genmo2024mochi,

      title={Mochi 1},

      author={Genmo Team},

      year={2024},

      publisher = {GitHub},

      journal = {GitHub repository},

      howpublished={\url{https://github.com/genmoai/models}}

}

@article{modify,

  doi = {10.5281/ZENODO.1207631},

  

  url = {https://zenodo.org/record/1207631},

  

  author = {Ha, David and Schmidhuber, Jürgen},

  

  title = {World Models},

  

  publisher = {Zenodo},

  

  year = {2018},

  

  copyright = {Creative Commons Attribution 4.0}

}

@misc{assran2025vjepa2selfsupervisedvideo,

      title={V-JEPA 2: Self-Supervised Video Models Enable Understanding, Prediction and Planning}, 

      author={Mido Assran and Adrien Bardes and David Fan and Quentin Garrido and Russell Howes and Mojtaba and Komeili and Matthew Muckley and Ammar Rizvi and Claire Roberts and Koustuv Sinha and Artem Zholus and Sergio Arnaud and Abha Gejji and Ada Martin and Francois Robert Hogan and Daniel Dugas and Piotr Bojanowski and Vasil Khalidov and Patrick Labatut and Francisco Massa and Marc Szafraniec and Kapil Krishnakumar and Yong Li and Xiaodong Ma and Sarath Chandar and Franziska Meier and Yann LeCun and Michael Rabbat and Nicolas Ballas},

      year={2025},

      eprint={2506.09985},

      archivePrefix={arXiv},

      primaryClass={cs.AI},

      url={https://arxiv.org/abs/2506.09985}, 

}

@misc{bruce2024geniegenerativeinteractiveenvironments,

      title={Genie: Generative Interactive Environments}, 

      author={Jake Bruce and Michael Dennis and Ashley Edwards and Jack Parker-Holder and Yuge Shi and Edward Hughes and Matthew Lai and Aditi Mavalankar and Richie Steigerwald and Chris Apps and Yusuf Aytar and Sarah Bechtle and Feryal Behbahani and Stephanie Chan and Nicolas Heess and Lucy Gonzalez and Simon Osindero and Sherjil Ozair and Scott Reed and Jingwei Zhang and Konrad Zolna and Jeff Clune and Nando de Freitas and Satinder Singh and Tim Rocktäschel},

      year={2024},

      eprint={2402.15391},

      archivePrefix={arXiv},

      primaryClass={cs.LG},

      url={https://arxiv.org/abs/2402.15391}, 

}

@article{parkerholder2024genie2,

  title         = {Genie 2: A Large-Scale Foundation World Model},

  author        = {Jack Parker-Holder and Philip Ball and Jake Bruce and Vibhavari Dasagi and Kristian Holsheimer and Christos Kaplanis and Alexandre Moufarek and Guy Scully and Jeremy Shar and Jimmy Shi and Stephen Spencer and Jessica Yung and Michael Dennis and Sultan Kenjeyev and Shangbang Long and Vlad Mnih and Harris Chan and Maxime Gazeau and Bonnie Li and Fabio Pardo and Luyu Wang and Lei Zhang and Frederic Besse and Tim Harley and Anna Mitenkova and Jane Wang and Jeff Clune and Demis Hassabis and Raia Hadsell and Adrian Bolton and Satinder Singh and Tim Rockt{\"a}schel},

  year          = {2024},

  url           = {https://deepmind.google/discover/blog/genie-2-a-large-scale-foundation-world-model/}

}

@misc{wu2025geometryforcingmarryingvideo,

      title={Geometry Forcing: Marrying Video Diffusion and 3D Representation for Consistent World Modeling}, 

      author={Haoyu Wu and Diankun Wu and Tianyu He and Junliang Guo and Yang Ye and Yueqi Duan and Jiang Bian},

      year={2025},

      eprint={2507.07982},

      archivePrefix={arXiv},

      primaryClass={cs.CV},

      url={https://arxiv.org/abs/2507.07982}, 

}

@article{lin2025uniworld,

  title={UniWorld: High-Resolution Semantic Encoders for Unified Visual Understanding and Generation},

  author={Lin, Bin and Li, Zongjian and Cheng, Xinhua and Niu, Yuwei and Ye, Yang and He, Xianyi and Yuan, Shenghai and Yu, Wangbo and Wang, Shaodong and Ge, Yunyang and others},

  journal={arXiv preprint arXiv:2506.03147},

  year={2025}

}
\bibliographystyle{abbrv}

%%%%%%%%%%%%%%%%%%%%%%%%%%%%%%%%%%%%%%%%%%%%%%%%%%%%%%%%%%%%

\appendix

\section{Motivation Experiments}
\label{app:A}
\setcounter{table}{0} 
\renewcommand{\thetable}{A.\arabic{table}}

Investigating the feasibility of simultaneously assimilating comprehensive physical and semantic knowledge, we implemented an extended experimental scheme labeled \textbf{Extend-VideoREPA}, incorporating three independent projection layers—configured with output dimensions of 768, 2048, and 768—to map the VAE latent into the distinct embedding spaces of \textbf{DINOv2}, \textbf{VGGT}, and \textbf{VideoMAEv2}. 

By formulating the total training objective as the summation of the standard diffusion loss and the independent Token Relation Distillation (TRD) terms from each expert branch, we compel the generative backbone to concurrently reconstruct the internal relational topology of these diverse feature spaces, thereby internalizing multi-source knowledge through a unified soft alignment paradigm.

\section{Implementation Details}
\label{app:B}
\setcounter{table}{0} 
\renewcommand{\thetable}{B.\arabic{table}}

\paragraph{Training Configuration.}
We initialize the generative backbone with the pre-trained weights of Wan2.1, while the newly introduced expanded projection layers are initialized to zero to strictly preserve the initial generative distribution. We employ the AdamW optimizer with hyperparameters $\beta_1=0.9$, $\beta_2=0.99$, and a weight decay of $\lambda=0.2$. The learning rate is set to $1e-5$ with seed $42$, utilizing a linear warmup for the first 400 steps followed by a consistency-constrained annealing strategy.To optimize memory efficiency and throughput, we utilize Mixed Precision Training (BF16) alongside gradient checkpointing. 

\paragraph{Feature Extraction and Alignment.}
For the World Knowledge priors, we employ visual foundation models to extract offline features, ensuring high-quality guidance signals.
Specifically, dense optical flow fields are first computed using RAFT~\cite{teed2020raftrecurrentallpairsfield}. Crucially, to align with the generative backbone, these 2D displacement fields are converted into RGB visualizations (as detailed in Section~\ref{subsec:preprocessing}) and subsequently encoded by the frozen Wan2.1 VAE~\cite{wan2025wanopenadvancedlargescale}. This maps the motion dynamics into the native latent space, resulting in a temporal feature dimension of $C_{temporal}=16$.
For semantic and spatial representations, we utilize the frozen DINOv2~\cite{oquab2024dinov2learningrobustvisual} and VGGT~\cite{wang2025vggtvisualgeometrygrounded} encoders. After spatial alignment and temporal pooling, we apply PCA to compress the channel dimensions of these features to $C_{semantic}=8$ and $C_{spatial}=8$, respectively.
Consequently, the final composite input fed into the expanded projection layer consists of the noisy video latents $C_{vae}=16$ concatenated with the WorldKnowledge priors, yielding a total input channel dimensionality of $C_{total} =48$.

\paragraph{Inference Setup.}
During inference, videos are generated using the Flow Matching Euler Discrete Scheduler with 50 denoising steps. We apply the proposed Multi-Faceted Self-Guidance with a text guidance scale of $w_{text}=5$ to ensure prompt fidelity. The auxiliary guidance scales for motion, semantics, and geometry are empirically set to $w_{temporal}=w_{semantic}=w_{spatial}=1$. This configuration was chosen based on ablation studies to provide subtle yet effective structural rectification without introducing high-frequency artifacts.

\paragraph{Inference Cost Analysis}
Integrating World Knowledge introduces computational overhead during inference. Specifically, the joint latent space increases peak VRAM by approximately 10.3\%. Additionally, inference time doubles from 2 to 4 minutes per video. This latency stems from the Multi-Source Inner-Guidance (Section~\ref{subsec:guidance}), requiring additional forward passes through the network. Despite the increased computational and memory footprints, the substantial improvements in trajectory control and generation quality justify these trade-offs.

\section{Detailed Quantitative Comparison}
\label{app:C}
\setcounter{table}{0} 
\renewcommand{\thetable}{C.\arabic{table}}
\subsection{Results of WorldScore}
Detailed multidimensional metrics for WorldScore are presented in Table\ref{tab:worldscore-static} and Table\ref{tab:worldscore-dynamic}.
\begin{table}[h]
    \centering
    \small
    \caption{Quantitative comparison on WorldScore (Static Attributes). The best and \underline{second-best} results are highlighted in \textbf{bold} and \underline{underlined}.}
    \label{tab:worldscore-static}
    \setlength{\tabcolsep}{2pt} 
    \begin{tabular}{lccccccc}
        \toprule
        \multirow{2}{*}[2.5ex]{\textbf{ Method}} & 
        \textbf{\shortstack{Camera\\Control}} & 
        \textbf{\shortstack{Object\\Control}} & 
        \textbf{\shortstack{Content\\Alignment}} & 
        \textbf{\shortstack{3D\\Consistency}} & 
        \textbf{\shortstack{Photometric\\Consistency}} & 
        \textbf{\shortstack{Style\\Consistency}} & 
        \textbf{\shortstack{Subjective\\Quality}} \\
        \midrule
        Wan2.1-T2V-1.3B & \underline{23.50} & \underline{52.34} & \textbf{66.83} & \underline{72.31} & 74.40 & 34.08 & 52.21 \\
        Wan2.1-T2V-1.3B(FT.) & \textbf{23.66} & 49.66 & \underline{65.35} & 71.73 & 74.62 & \textbf{39.04} & \textbf{53.83} \\
        VideoJAM & 22.48 & \textbf{57.34} & 59.32 & 71.53 & \underline{75.39} & 25.90 & 44.03 \\
        \textbf{DreamWorld (Ours)} & 22.81 & 51.94 & 64.95 & \textbf{73.16} & \textbf{77.55} & \underline{35.75} & \underline{52.45} \\
        \bottomrule
    \end{tabular}
    % \vspace{-10pt}
\end{table}

\begin{table}[h]
    \centering
    \small
    \caption{Quantitative comparison on WorldScore (Dynamic Attributes, Summary and Overall Score).The best and \underline{second-best} results are highlighted in \textbf{bold} and \underline{underlined}.}
    \label{tab:worldscore-dynamic}
    \setlength{\tabcolsep}{7.4pt} 
    \begin{tabular}{lcccccc}
        \toprule
        \multirow{2}{*}[-2ex]{\textbf{ Method}} & 
        \multicolumn{3}{c}{\textbf{\small Dynamic Attributes}} & 
        \multicolumn{2}{c}{\textbf{\small Summary}} & 
        \multirow{2}{*}[-2ex]{\textbf{\small \shortstack{ Overall\\Score}}} \\
        \cmidrule(lr){2-4} \cmidrule(lr){5-6} 
        & \textbf{\shortstack{Motion\\Accuracy}} & \textbf{\shortstack{Motion\\Magnitude}} & \textbf{\shortstack{Motion\\Smoothness}}
        & \textbf{\shortstack{Static\\Score}} & \textbf{\shortstack{Dynamic\\Score}} & \\ 
        \midrule
        Wan2.1-T2V-1.3B & 32.36 & \textbf{34.14} & 68.18 & 53.66 & 44.89 & \underline{51.03} \\
        Wan2.1-T2V-1.3B(FT.) & \textbf{35.49} & 31.35 & 64.76 & \underline{53.98} & 43.86 & 50.95 \\
        VideoJAM & 33.38 & \underline{32.51} & \textbf{71.95} & 50.86 & \textbf{45.94} & 49.38 \\
        \textbf{DreamWorld (Ours)} & \underline{34.75} & 32.44 & \underline{69.02} & \textbf{54.09} & \underline{45.41} & \textbf{51.48} \\
        \bottomrule
    \end{tabular}
    \vspace{-10pt}
\end{table}
\begin{table}[tb!]
    \centering
    \scriptsize  
    
    \caption{\textbf{Results of VBench} (Part 1/2): Consistency, Motion, and Object-level Metrics.}
    \label{tab:vbench-detail-1}
    \setlength{\tabcolsep}{2.1pt} 
    \begin{tabular}{lcccccccccc}
        \toprule
        \multirow{2}{*}[2ex]{\textbf{Method}} 
        & \textbf{\shortstack{Subject\\Consistency}} 
        & \textbf{\shortstack{Background\\Consistency}}
        & \textbf{\shortstack{Temporal\\Flickering}} 
        & \textbf{\shortstack{Motion\\Smoothness}}
        & \textbf{\shortstack{Dynamic\\Degree}} 
        & \textbf{\shortstack{Aesthetic\\Quality}}
        & \textbf{\shortstack{Imaging\\Quality}} 
        & \textbf{\shortstack{Object\\Class}}
        & \textbf{\shortstack{Multiple\\Objects}}
        & \textbf{\shortstack{Human\\Action}}\\
        \midrule
        Wan2.1-T2V-1.3B & 91.83 & 94.71 & 99.17 & 96.51 & 65.00 & 57.24 & 59.14 & 76.09 & 53.66 & 74.60 \\
        Wan2.1-T2V-1.3B(FT.) & 93.59 & 95.81 & \textbf{99.36} & 97.21 & 54.08 & \textbf{61.13} & 63.27 & 79.90 & 58.20 & 78.98 \\
        VideoJAM & 91.51 & \textbf{96.01} & 99.13 & 96.05 & 73.88 & 58.41 & 62.68 & 79.22 & 59.66 & 79.20 \\
        \textbf{DreamWorld (Ours)} & \textbf{93.62} & 94.95 & 98.81 & \textbf{98.07} & \textbf{79.16} & 59.59 & \textbf{66.76} & \textbf{81.32} & \textbf{65.03} & \textbf{81.20} \\
        \bottomrule
    \end{tabular}

    \vspace{0.6em} 

    \caption{\textbf{Results of VBench} (Part 2/2): Appearance, Style, and Overall Scores.}
    \label{tab:vbench-detail-2}
    \setlength{\tabcolsep}{4.4pt} 
    \begin{tabular}{lccccccccc}
        \toprule
        \multirow{2}{*}[2.5ex]{\textbf{Method}} & 
        \multirow{2}{*}[2.5ex]{\textbf{Color}} & 
        \textbf{\shortstack{Spatial\\Relationship}} &
        \multirow{2}{*}[2.5ex]{\textbf{Scene}} &  
        \textbf{\shortstack{Appearance\\Style}} & 
        \textbf{\shortstack{Temporal\\Style}} & 
        \textbf{\shortstack{Overall\\Consistency}} & 
        \textbf{\shortstack{Quality\\Score}} & 
        \textbf{\shortstack{Semantic\\Score}} & 
        \textbf{\shortstack{Overall\\Score}} \\
        \midrule
        Wan2.1-T2V-1.3B & 89.93 & 62.37 & 20.03 & \textbf{22.91} & 22.23 & 24.25 & 79.81 & 65.43 & 76.93 \\
        Wan2.1-T2V-1.3B(FT.) & 88.57 & 63.31 & 28.55 & 22.38 & 23.93 & 24.94 & 81.26 & 68.47 & 78.71 \\
        VideoJAM & 88.92 & 66.17 & 28.34 & 22.67 & \textbf{24.03} & \textbf{25.01} & 81.18 & 69.08 & 78.76 \\
        \textbf{DreamWorld (Ours)} & \textbf{92.61} & \textbf{70.47} & \textbf{29.71} & 22.14 & 23.82 & 24.87 & \textbf{83.49} & \textbf{70.89} & \textbf{80.97} \\
        \bottomrule
    \end{tabular}
    \vspace{-10pt}
\end{table}

\begin{table}[tb!]
    \centering
    \scriptsize 

    \caption{\textbf{VBench-2.0 results} (part 1/3).}
    \label{tab:vbench2_part1}
    \setlength{\tabcolsep}{3.4pt} 
    \begin{tabular}{lcccccccc}
        \toprule
        \multirow{2}{*}[2.5ex]{\textbf{Method}} & 
        \textbf{ \shortstack{Human\\Identity}} & 
        \textbf{ \shortstack{Dynamic Spatial\\Relationship}} & 
        \textbf{ \shortstack{Complex\\Landscape}} & 
        \textbf{ \shortstack{Instance\\Preservation}} & 
        \textbf{ \shortstack{Multi-View\\Consistency}} & 
        \textbf{ \shortstack{Human\\Clothes}} & 
        \textbf{ \shortstack{Dynamic\\Attribute}} & 
        \textbf{ \shortstack{Complex\\Plot}} \\
        \midrule
        Wan2.1-T2V-1.3B      & 64.21 & 26.57 & 16.22 & 80.41 & \textbf{14.63} & \textbf{99.55} & 22.34 & 10.53 \\
        Wan2.1-T2V-1.3B (FT.) & \textbf{72.43} & \textbf{28.50} & \textbf{21.66} & 88.24 & 8.24  & 90.13 & 22.71 & \textbf{11.77} \\
        VideoJAM             & 72.14 & 24.08 & 15.66 & \textbf{90.41} & 6.79  & 97.74 & 24.37 & 10.33 \\
        \textbf{DreamWorld (Ours)} & 71.68 & 24.15 & 14.66 & 86.14 & 12.41 & 99.11 & \textbf{26.73} & 10.44 \\
        \bottomrule
    \end{tabular}
    
    \vspace{0.6em} 

    \caption{\textbf{VBench-2.0 results} (part 2/3).}
    \label{tab:vbench2_part2}
    \setlength{\tabcolsep}{3.2pt}
    \begin{tabular}{lcccccccc}
        \toprule
        \multirow{2}{*}[2.5ex]{\textbf{Method}} & 
        \multirow{2}{*}[2.5ex]{\textbf{ Mechanics}} & 
        \textbf{ \shortstack{Human\\Anatomy}} & 
        \multirow{2}{*}[2.5ex]{\textbf{ Composition}} & 
        \textbf{ \shortstack{Human\\Interaction}} & 
        \textbf{ \shortstack{Motion\\Rationality}} & 
        \multirow{2}{*}[2.5ex]{\textbf{ Material}} & 
        \multirow{2}{*}[2.5ex]{\textbf{ Diversity}} & 
        \textbf{ \shortstack{Motion Order\\Understanding}} \\
        \midrule
        Wan2.1-T2V-1.3B      & 85.71 & 64.51 & 40.19 & 15.00 & 37.93 & 57.77 & 51.65 & 7.31 \\
        Wan2.1-T2V-1.3B (FT.) & \textbf{88.63} & 68.71 & 39.35 & 14.66 & 37.35 & \textbf{60.67} & 46.91 & 6.42 \\
        VideoJAM             & 80.31 & 66.51 & 44.81 & 17.33 & \textbf{39.37} & 56.62 & 53.84 & 7.53 \\
        \textbf{DreamWorld (Ours)} & 86.30 & \textbf{69.35} & \textbf{45.01} & \textbf{19.00} & 38.51 & 55.17 & \textbf{56.76} & \textbf{8.88} \\
        \bottomrule
    \end{tabular}
    
    \vspace{0.6em}

    \caption{\textbf{VBench-2.0 results} (part 3/3).}
    \label{tab:vbench2_part3}
    \setlength{\tabcolsep}{3.2pt}
    \begin{tabular}{lcccccccc}
        \toprule
        \multirow{2}{*}[2.5ex]{\textbf{Method}} & 
        \textbf{ \shortstack{Camera\\Motion}} & 
        \multirow{2}{*}[2.5ex]{\textbf{ Thermotics}} & 
        \textbf{ \shortstack{Creativity\\Score}} & 
        \textbf{ \shortstack{Commonsense\\Score}} & 
        \textbf{ \shortstack{Controllability\\Score}} & 
        \textbf{ \shortstack{Human Fidelity\\Score}} & 
        \textbf{ \shortstack{Physics\\Score}} & 
        \textbf{ \shortstack{Total\\Score}} \\
        \midrule
        Wan2.1-T2V-1.3B      & 19.75 & 65.27 & 45.92 & 59.17 & 16.81 & 76.09 & \textbf{55.85} & 50.77 \\
        Wan2.1-T2V-1.3B (FT.) & \textbf{23.14} & 60.49 & 43.13 & 62.80 & \textbf{18.41} & 77.09 & 54.51 & 51.18 \\
        VideoJAM             & 15.04 & 61.97 & 49.32 & \textbf{64.89} & 16.33 & 78.68 & 52.42 & 52.33 \\
        \textbf{DreamWorld (Ours)} & 14.81 & \textbf{66.43} & \textbf{50.89} & 61.82 & 16.95 & \textbf{80.11} & 55.07 & \textbf{52.97} \\
        \bottomrule
    \end{tabular}
    \vspace{-10pt}
\end{table}

\subsection{VBench and VBench2.0 Results}
We report the detailed breakdown of metrics on both VBench and VBench-2.0 benchmarks, providing a granular analysis of model performance. The comprehensive evaluations for VBench are presented in Table~\ref{tab:vbench-detail-1} and Table~\ref{tab:vbench-detail-2}, while the VBench-2.0 results are detailed in Table~\ref{tab:vbench2_part1}, Table~\ref{tab:vbench2_part2}, and Table~\ref{tab:vbench2_part3}.

As evidenced by the quantitative data, \textbf{DreamWorld} demonstrates robust superiority over the baselines. In VBench, our method achieves significant margins in Spatial Relationship and Dynamic Degree, which contributes to a leading Overall Score of 80.97. Furthermore, on the more challenging VBench-2.0, DreamWorld secures the highest Total Score (52.97) and excels in complex categories such as Human Fidelity and Motion Order Understanding.  These results validate that our joint world modeling paradigm effectively internalizes multi-modal priors, resulting in generated content with real-world fidelity.

\section{More Ablation Study}
\label{app:E}
\setcounter{table}{0} 
\renewcommand{\thetable}{E.\arabic{table}}
\begin{figure}[tb!]
    \centering
    \begin{minipage}{0.48\linewidth}
        \centering
        \includegraphics[width=\linewidth]{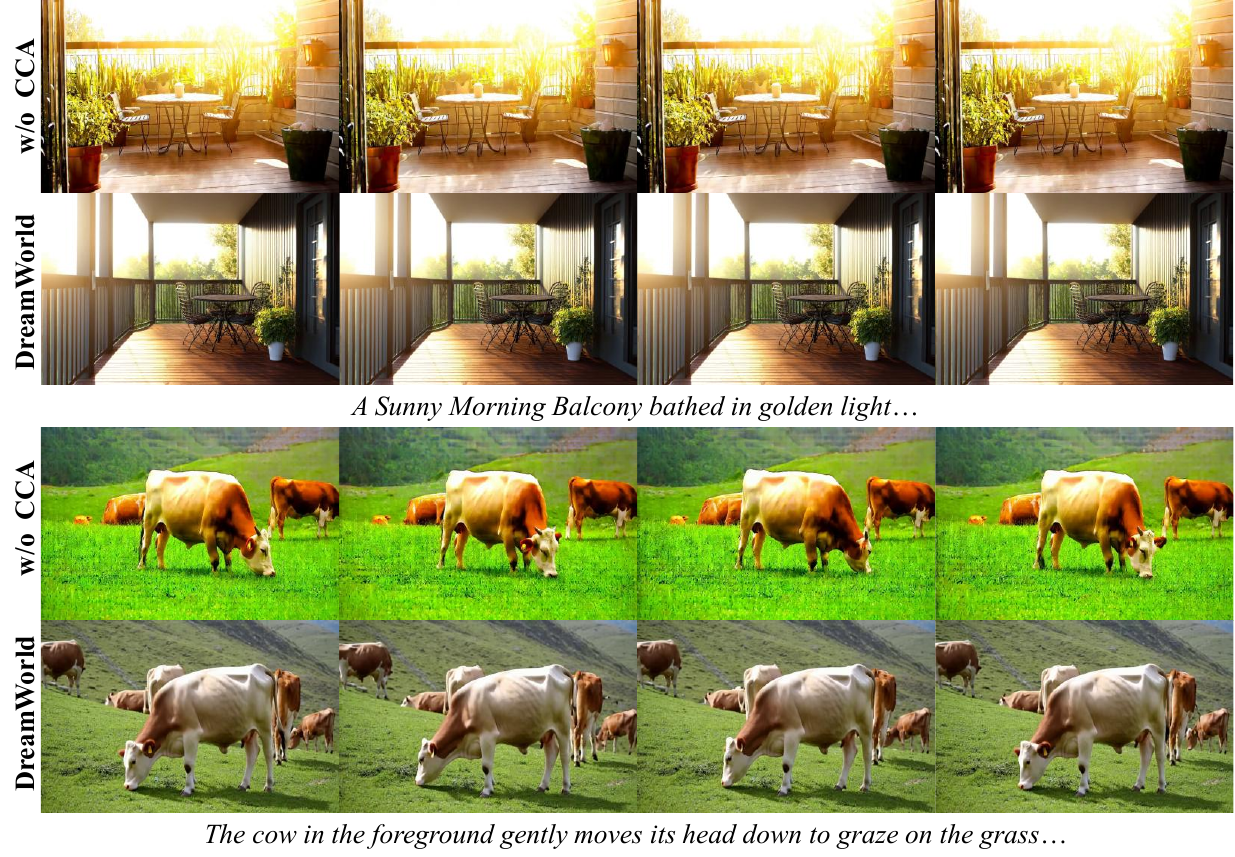}
        \caption{\textbf{Effectiveness of CCA.} Without CCA, the generated videos suffer from severe flickering and abnormal highlighting artifacts, leading to degraded visual quality.}
        \label{fig:app_cca_ablation}
    \end{minipage}
    \hfill
    \begin{minipage}{0.48\linewidth}
        \centering
        \includegraphics[width=\linewidth]{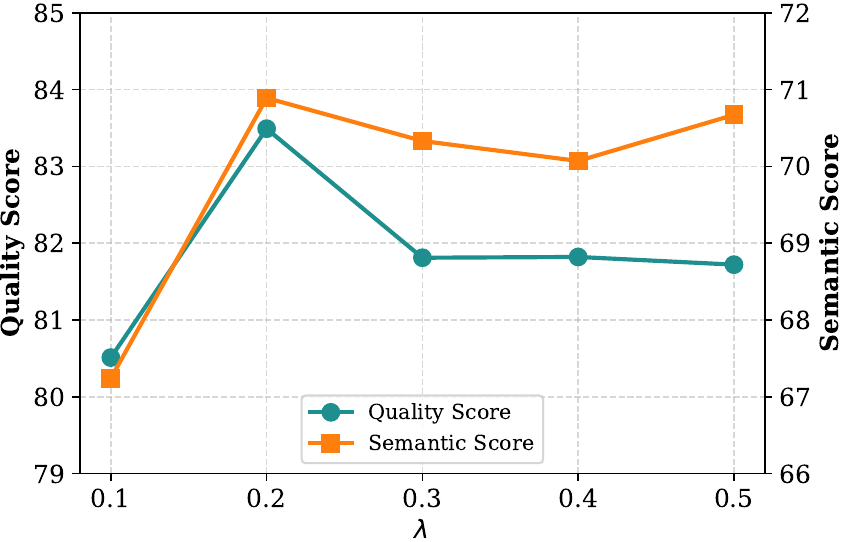}
        \caption{\textbf{Influence of loss weight $\lambda$.} Quantitative results show that $\lambda=0.2$ achieves the best trade-off between generation quality and semantic alignment.}
        \label{fig:app_lambda_metrics}
    \end{minipage}
\end{figure}

\begin{table*}[tb!]
    \centering
    \begin{minipage}[t]{0.48\linewidth}
        \centering
        \caption{\textbf{Quantitative ablation of the CCA schedule.}
        We compare different strategies for scheduling the world loss weight.}
        % \vspace{6pt}
        \label{tab:app_cca_schedule}
        \setlength{\tabcolsep}{4pt}
        \resizebox{\linewidth}{!}{%
        \renewcommand{\arraystretch}{1.48}
        \begin{tabular}{lccc}
            \toprule
            \textbf{Configuration} & \textbf{Quality} & \textbf{Semantic} & \textbf{Overall} \\
            \midrule
            No World Loss (Baseline) & 81.26 & 68.47 & 78.71 \\
            Fixed World-Loss Weight & 80.76 & 67.85 & 78.19 \\
            Increasing Schedule & 81.13 & 68.02 & 78.51 \\
            \midrule
            \textbf{Decreasing Schedule (Ours)}
                & \textbf{83.49} & \textbf{70.89} & \textbf{80.97} \\
            \bottomrule
        \end{tabular}%
        }
    \end{minipage}
    \hfill
    \begin{minipage}[t]{0.48\linewidth}
        \centering
        \caption{\textbf{Comparison of guidance configurations.}
        Different guidance configurations are compared for cost and quality.}
        % \vspace{6pt}
        \label{tab:app_guidance_tradeoff}
        \setlength{\tabcolsep}{4pt}
        \resizebox{\linewidth}{!}{%
        \renewcommand{\arraystretch}{1.03}
        \begin{tabular}{lccc}
            \toprule
            \textbf{Configuration} & \textbf{Latency} & \textbf{VRAM} & \textbf{Overall} \\
            \midrule
            No Guidance & 48 & 25.55 & 68.84 \\
            Text Guidance Only & 100 & 26.55 & 78.13 \\
            DINOv2 & 144 & 28.03 & 78.82 \\
            DINOv2 + VGGT & 193 & 30.73 & 79.71 \\
            \midrule
            \textbf{Multi-Source (Ours)}
                & 240 & 33.42 & \textbf{80.97} \\
            \bottomrule
        \end{tabular}%
        }
    \end{minipage}
\end{table*}

\paragraph{Effectiveness of CCA.}
Further ablation studies demonstrating the significant impact of the CCA mechanism on overall model performance are visually presented in Figure\ref{fig:app_cca_ablation}.To further quantitatively investigate the effect of the CCA schedule, we compare different world-loss weighting strategies, including a fixed weight, an increasing schedule, and our decreasing schedule. As shown in Table~\ref{tab:app_cca_schedule}, the decreasing schedule consistently achieves the best performance across all three metrics. This result suggests that applying stronger structural constraints at earlier training stages while gradually relaxing them at later stages provides a more effective balance between world knowledge learning and visual generation quality.

\paragraph{Guidance Efficiency.}
We further investigate the computational overhead and performance gains of different guidance configurations. Specifically, the single World-Guidance branch uses only DINOv2 features, while the two World-Guidance branches jointly use DINOv2 and VGGT features. As shown in Table~\ref{tab:app_guidance_tradeoff}, introducing additional guidance sources improves generation quality at the cost of increased inference latency and GPU memory consumption. Our Multi-Source Guidance achieves the highest VBench overall score of $80.97$, compared with $78.82$ using DINOv2 alone and $79.71$ using DINOv2 and VGGT.

\paragraph{Influence of $\lambda$.}
We investigate the sensitivity of our model to the  loss weight $\lambda$, which controls the strength of the world knowledge priors. To quantify the trade-off between generative fidelity and condition adherence, we report both the Quality Score and Semantic Score on VBench across varying $\lambda$ values. As illustrated in Figure.~\ref{fig:app_lambda_metrics}, the performance of both metrics peaks at $\lambda=0.2$. We observe that lower values fail to effectively enforce the physical and semantic constraints, whereas excessive weights interfere with the diffusion backbone's distribution, leading to a degradation in visual quality. Consequently, we adopt $\lambda=0.2$ as the optimal configuration for our full model.

\section{Broader Impacts}
\label{app:F}
Our DreamWorld demonstrates a marked improvement over existing video generation baselines, establishing new state-of-the-art benchmarks in world consistency and physical commonsense. This enhancement is pivotal in advancing general-purpose world simulators and high-fidelity video synthesis technologies. While our contributions are essentially within the scope of methodology and fundamental theory, we are fully aware of the ethical challenges facing generative media, such as the potential misuse for creating deceptive deepfakes. Although this work does not introduce new specific ethical risks beyond those inherent in existing video synthesis techniques, we recognize that highly realistic generation can be exploited for disinformation. We share the position of the broader academic community in firmly supporting responsible and sustainable technology development to ensure the safety and fairness of generative tools, including the adoption of appropriate mitigation strategies.

\section{Safeguards}
\label{app:G}
To mitigate the potential dual-use risks and misuse of our video generation architecture, we release the model and code under a strict Acceptable Use Policy (AUP). Users who wish to access and utilize our model are required to explicitly agree to this license, which strictly prohibits the generation of harmful, deceptive, or non-consensual content. This licensing approach serves as a primary safeguard to promote responsible downstream usage while still supporting the open research community.

\section{More qualitative results}
\label{app:D}
\begin{figure}[p]
    \centering
    \includegraphics[width=1.0\linewidth]{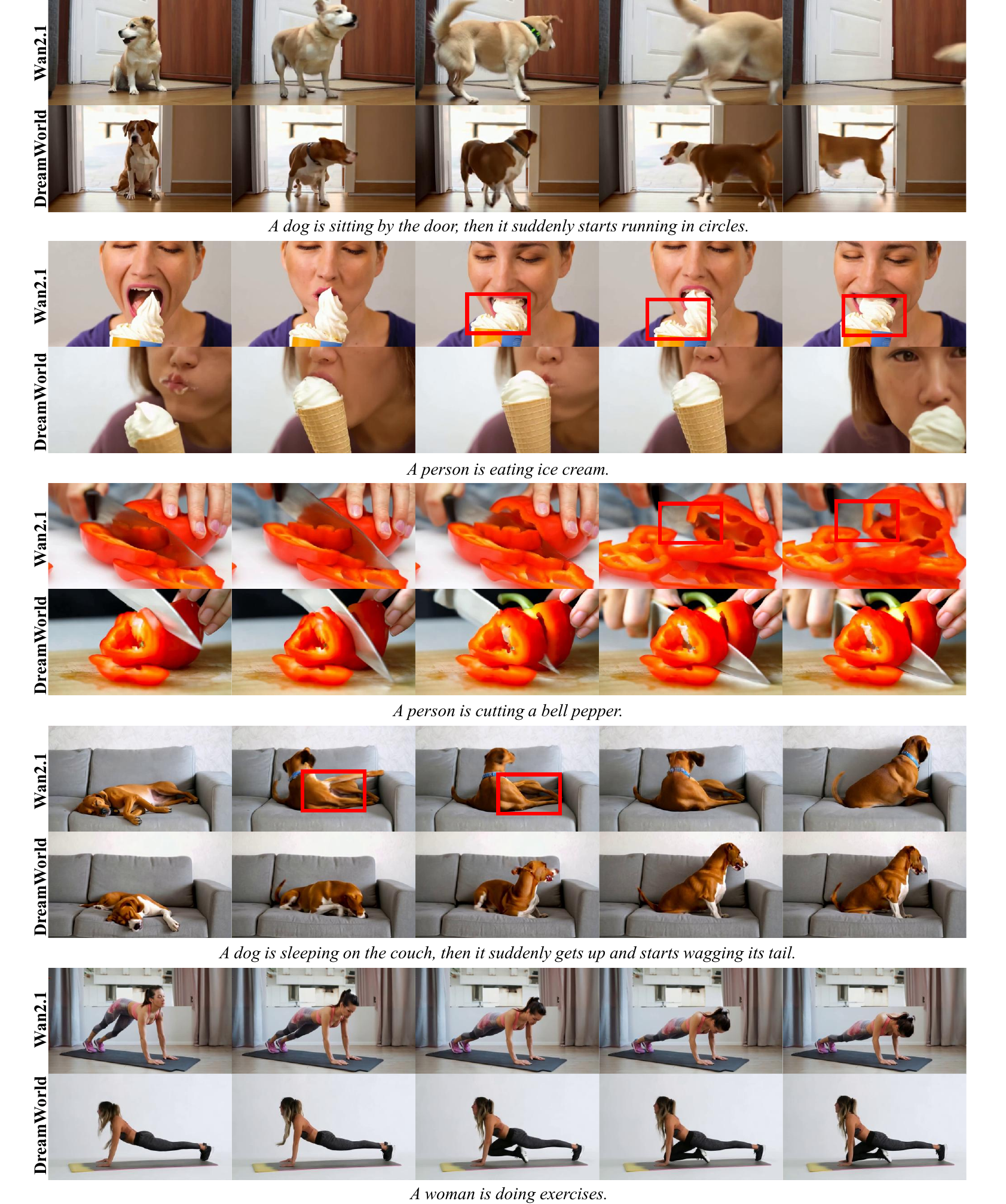}
    \caption{\textbf{Additional qualitative comparisons.}The red rectangles highlight anomalies(e.g., object penetration, unnatural disappearance, and limb distortion).}
    \label{fig:app_qualitative}
\end{figure}

We provide additional qualitative comparisons in Figure~\ref{fig:app_qualitative} to further substantiate the superiority of DreamWorld over the baseline Wan2.1. These visual results underscore our model's robustness across semantic, spatial, and temporal dimensions. Specifically, The baseline frequently succumbs to semantic hallucinations and spatial conflicts, where objects unnaturally fuse or distort. Conversely, DreamWorld maintains clear object boundaries and logical interaction physics. 

\clearpage
\clearpage
%%%%%%%%%%%%%%%%%%%%%%%%%%%%%%%%%%%%%%%%%%%%%%%%%%%%%%%%%%%%

\end{document}